\documentclass{article} 
\usepackage{nips15submit_e,times}
\usepackage{hyperref}
\usepackage{url}
\usepackage{amsmath}
\usepackage{algorithm}
\usepackage[noend]{algpseudocode}
\usepackage{graphicx}
\usepackage{caption}
\usepackage{subcaption}
\usepackage{epstopdf}

\usepackage{float}
\expandafter\def\expandafter\normalsize\expandafter{%
	\normalsize
	\setlength\abovedisplayskip{4pt}
	\setlength\belowdisplayskip{4pt}
	\setlength\abovedisplayshortskip{4pt}
	\setlength\belowdisplayshortskip{4pt}
}
\title{ Gaussian Process for Noisy Inputs with Ordering Constraints}

\author{
	Cuong Tran\\
	Department of Computer Science\\
	Rutgers University\\
	\texttt{cuong.tran@cs.rutgers.edu} \\
	\and
	\textbf{Vladimir Pavlovic }\\
	Department of Computer Science \\
	Rutgers University\\
	\texttt{vladimir@cs.rutgers.edu} 
		\AND
		Robert Kopp \\
		 Department of Earth \& Planetary Sciences \\
		Rutgers University\\
		\texttt{robert.kopp@rutgers.edu} \\
}

\nipsfinalcopy 

\begin{document}

	\maketitle
	
	\begin{abstract}
		We study the Gaussian Process regression model in the context of training data with noise in both input and output. The presence of two sources of noise makes the task of learning accurate predictive models extremely challenging.  However, in some instances additional constraints may be available that can reduce the uncertainty in the resulting predictive models.  In particular, we consider the case of monotonically ordered latent input, which occurs in many application domains that deal with temporal data.  We present a novel inference and learning approach based on non-parametric Gaussian variational approximation to learn the GP model while taking into account the new constraints. The resulting strategy allows one to gain access to posterior estimates of both the input and the output and results in improved predictive performance.  We compare our proposed models to state-of-the-art Noisy Input Gaussian Process (NIGP) and other competing approaches on synthetic and real sea-level rise data.  Experimental results suggest that the proposed approach consistently outperforms selected methods while, at the same time, reducing the computational costs of learning and inference. 
	\end{abstract}

	\section{Introduction}
	Uncertain or noisy data, both in input and the output, is a common problem that cannot be avoided in many real world applications. Neglecting this uncertainty will results in inaccurate predictive models, particularly when the noise is large.  Most machine learning models and settings only consider the output noise, and devise ways to effectively mitigate its presence.  Noise in the input is considered less frequently, typically in the context of error-in-variable models \cite{dellaportas:stephens:95} The presence of input noise is typically more difficult to handle than the additive output noise, largely because of the nonlinear dependence of predictors on its input.  To address this challenge traditional sampling-based Monte Carlo Markov chain (MCMC) techniques are often employed, however they are time-consuming and will not be appropriate for large datasets. Explicit integrating out of the input uncertainty, when the input density is known, is intractable in common situations \cite{NIPS2011_4295} , \cite{titsias2010bayesian}. 
	
	The challenge of handling noisy input becomes more daunting when other sources of prior knowledge of the unobserved true input are present and need to be taken into account.  In many applications dealing with time-series data and in particular in earth sciences, it may be known that the samples $\{(y_i,t_i)\}$, with uncertainty in both the output $y_i$ and input $t_i$, must be ordered, e.g., that the underlying latent noise-free estimates $\tau_i \leq \tau_{i+1}$ corresponding to $t_i, t_{i+1}$.  For instance, in measurements of historical sea-level, which are often based on geological records, it is known that certain measurements precede others in time, although their exact ages remain unknown.  The uncertainty in input (age) obtained from carbon $^{14}C$-dating can often be large enough to yield high likelihood of miss-ordering e.g., $Pr( t_i > t_{i+1} ) > 1-\epsilon$, $\epsilon>0$, yet requires $\tau_i \leq \tau_{i+1}$.  Incorporating such ordering constraints into noisy input learning is, however, nontrivial.  
	
	A number of approaches to dealing with input noise have been developed in the context of Gaussian Processes (GPs).  For example, Girard and Smith in \cite{girard2003learning} proposed to use a second Taylor expansion around the input mean to obtain a new corrected GP that accounts for the uncertain inputs. An alternative approach is to correct the covariance matrix in GPs under the presence of input noise and was introduced in \cite{conf/iconip/DallaireBC09}. The corrected covariance matrix was determined by computing the expectation of the covariance function with respect of the input distributions.  The closed form of the expectation was provided in \cite{girard2003gaussian} for linear, polynomial and squared exponential covariance function. Recently,  McHutchon in \cite{Rasmussen:2005:GPM:1162254} developed a simple but effective method called noisy input GP (NIGP) that was showed to outperform the previous approaches. The basic idea of NIGP is to refer the input noise to the output noise by using a first order Taylor expansion around the noisy inputs, similar to traditional error-in-variable approaches. A procedure to iteratively optimize the input noise parameters and GP hyper-parameters was also provided. Although NIGP was shown to perform well on synthetic datasets, it remains to share common limitations with related approaches. In particular, incorporating prior information into the NIGP framework is challenging.  Next, NIGP may not perform well in cases of large input noise due to its dependence on (first order) Taylor expansion. Finally, NIGP does not provide an immediate means to estimating the posterior density of the latent input, a task which is often of interest in practical applications.
	
	In this work we propose a new approach to learning GP models from data corrupted by dual input-output noise, in the setting when ordering constraints on the latent input are present.  Depending on the quantification of ordering constraints, the task of learning the GP models, estimating the posterior of the latent (but ordered) input, and the posterior of the output become nontrivial.  In particular, the densities of interest cannot be computed analytically nor do they remain in the exponential family.  To address these challenges in Sec~\ref{npv} we formulate a non-parametric variational approach based on recent work in \cite{DBLP:journals/corr/abs-1206-4665} in the context of ordered input noisy GPs.  We demonstrate how additional approximations can be used to yield tractable inference and learning in these models, as outlined in Sec.~\ref{fast}.  Finally, in Sec.~\ref{exp} we demonstrate the utility of our approach by contrasting its performance to state-of-the-art models, including NIGP and a sampling-based MCMC solution.

	\section{Problem formulation}
	\label{headings}
	
	We consider the following non-linear regression model $y_i=f(\tau_i)+\epsilon_{y,i}$ where $\{\tau_i\}_{i=1}^N$ are explanatory variables , $y_i$ are response variables and $\epsilon_{y,i}$ are zero mean Gaussian output noise variables with known standard deviations $\sigma_{y,i}$. 
	In our work, the true input variables $\tau=\{ \tau_i\}_{i=1}^N $ are not observed and what we actually observe are their noisy versions. We assume a classical error-in-variable model here to obtain the noisy inputs: i.e $ t_i=\tau_i +\epsilon_{t,i} $, where $\epsilon_{t,i} \sim \mathcal{N} (\epsilon_{t,i} | 0, \sigma_{t,i})$ is an additive zero mean Gaussian noise that is independent from $\tau_i$.

	The latent true output variables $\{f(\tau_i)\}_{i=1}^{N},$ are assumed to have a GP prior with zero mean and  a covariance function $k_{\theta}(\tau,\tau' )$. We assume in this paper the covariance function is stationary. The reason that we use a GP framework here is because of its flexibility due to nonparametric property and its ability to handle uncertain data as previous works suggested. Learning in usual GPs involve choosing the optimal hyper-parameters  $\theta^{*}$  by maximizing the log-marginal likelihood: $ \theta^{*} =\arg \max_{\theta} \int \log \big( Pr(y|f) Pr(f|\tau)  \big) df =-y^T K_{\theta} y - log|K_{\theta}| +const $. Where $K_{\theta}$ is the training covariance matrix, $(K_{\theta})_{i,j} = k_{\theta}(\tau_i,\tau_j ) + \sigma_{y,i}^2 I(i-j) ; \forall i,j=1,2,..,n$ ; $I(.) $ is the indicator function.
	
	The prior knowledge in our model is that the latent true inputs $\tau_i$ satisfy: $\tau_{1}>\tau_{2}>...>\tau_{N}$. The final goal is to predict the function value $f(\tau^*)$ of an unseen sample $\tau^*$. Without noisy input data $t_i$,  $f(\tau^*)$ is Gaussian variable with mean and variance alternatively given by \cite{Rasmussen:2005:GPM:1162254}:

	\begin{equation}E[f(\tau^*)] =k(\tau, \tau^*)^T K^{-1}y \end{equation} 
	\begin{equation}Var[f(\tau^*] =k(\tau^* , \tau^*) -  k(\tau, \tau^*)K^{-1} k(\tau, \tau^*) \end{equation}

	\section{Nonparametric Gaussian variational inference model }
	\label{npv}
	
	Here we present our method that can  overcome those difficulties listed in previous sections. First, in order to guarantee the monotonic order of $\tau_i$ we use the following variable transformation:
	\begin{equation} r=\tau_n; l_i=\log(\tau_i-\tau_{i+1}) \  \forall i=1,2,..,n-1 \end{equation}
	
	We will model the random variables $r, l_i$ instead of $\tau_i$ because we do not have the constraints anymore. For simplicity, we assume that $l_i$ and $r$ have a uniform prior, i.e $Pr(l_i)\propto 1; \forall i; Pr(r) \propto 1$. The log mariginal likelihood $\log Pr(y | \theta)$ can be bounded below by introducing a variational distribution $Q(l,r)$ as follows:
	\begin{equation}
	\begin{aligned}
	\log Pr(y |\theta)=\log \bigg( \int \int Pr(y,l,r | \theta) dl dr \bigg) \geq  \int \int Q(l,r)\log \frac{Pr(y,l,r | \theta)}{Q(l,r)} dl dr
	\\= \int \int Q(l,r) \log \frac{Pr(l,r |y, \theta)Pr(y|\theta)}{Q(l,r)} dl dr  =\log Pr(y |\theta)- \int \int Q(l,r)\log \frac{Pr(l,r | y,\theta)}{Q(l,r)} dl dr
	\end{aligned}
	\end{equation}
	In order to maximize the log marginal likelihood, we seek to find a variational distribution $Q(l,r)$ which belongs to a tractable distribution family and minimize the KL divergence from $Q$ to  $Pr(l,r|y, \theta)$ .  We choose $Q$ to be a mixture of $K$ Gaussians to capture the possible multimodality of$Pr(l,r|y)$ \cite{DBLP:journals/corr/abs-1206-4665} 
	\begin{equation} Q( l,r | \Phi)=\frac{1}{K} \sum_{i=1}^{K} \mathcal{N}(l ,r |m_i, V_i )   \end{equation}
	\begin{equation}  V_i=diag(v_i );    \Phi=\{m_i, v_i\}; \forall i=1,2,..,K  \}\end{equation}
	
	Here we choose $V_i$ to be an isotropic covariance matrice for optimization convenience. The set of variational parameters $\Phi$ can be found by minimizing the KL divergence between the variational distribution $Q$ and the true posterior distribution $Pr(l,r| y,t, \sigma_t, \theta)$. The objective function that we need to minimize is: 
	\begin{equation}F(\Phi ; \theta)= KL\big[ Q(l, r)|| Pr( l,r |y, t, \sigma_t, \theta)    \big] =H[Q]-E_{Q}  \log Pr(t | l,r, \sigma_t ) -E_{Q} \log Pr(y | l; \theta) \end{equation}
	Where:
	
	\begin{equation} H[Q]=- \int \int Q(l,r) \log Q(l,r) dl dr  \end{equation}
	
	\begin{equation}E_{Q} \log Pr( t |l, r , \sigma_t)= E_{Q} \bigg( - \sum_{i=1}^{n-1} \frac{ \big(  r +\sum_{j=i}^{n-1} e^{l_i} -t_i\big)^2}{2 \sigma_{t,i}^2} - \frac{ (r-t_n)^2}{\sigma_{t,n}^2}\bigg) +const \end{equation}
	\begin{equation}E_{Q} \log Pr(y | l, \theta)= E_{Q}   \bigg( -\frac{1}{2}  y^T K(l)^{-1}y  -\frac{1}{2} \log|K(l)| + const \bigg) \end{equation}

	In the above equations, the entropy term $H[Q]$ can be bounded above by using Jensen inequality :
	\begin{equation} H[Q] \geq -\sum_{i=1}^K \frac{1}{K} \sum_{j=1}^K \mathcal{N}\big(m_i ;m_j, V_i +V_j \big)\end{equation}

	The  expected log likelihood with respect of $t$,  $ E_{Q}  \log Pr(t | l,r, \sigma_t ) $ can be computed analytically with detailed derivation included in the supplementary material..
	
	In Eq. (11) for stationary covariance functions the log marginal likelihood $\log Pr(y | l;\theta)$ depends only on $l$ but not $r$. Its expecation will be a highly nonlinear function of $l$ due to its appearance in the inverse matrix $K(l)$ and the expecation does not have a closed form.  We approximate this term by a second Taylor expansion around the means $ m_i , i=1,2,..,K$.
	
	\begin{equation}E_{Q} \log Pr(y | l; \theta) \approx \frac{1}{K} \sum_{i=1}^{K}  \bigg( \log Pr(y | m_i; \theta)   +\frac{1}{2} trace \big( \nabla^{2}_{l}   \log Pr(y | l; \theta)_{l=m_i} V_i \big)   \bigg)\end{equation} 
	
	We can iteratively optimize $\Phi$ and $\theta$ and based on the final optimal values of $m^{(l)}$ and $m^{(r)}$ to determine an estimation of the true inputs. Then we uses the these estimated quantities for future prediction based on Eq. (1) \& (2). One important point is that minimization of $F(\Phi ;\theta) $ requires us to compute the gradient and the Hessian of  $ \log Pr (y |l, \theta)$ as you can see in Eq. (11) which might take a lot of time. In the next section, we will present the key idea to compute these terms efficiently.
	
	\subsection{Using chain rule for faster computation }
	\label{fast}
	Procedures of computing the gradient  $\frac{\partial \log Pr (y |l) }{ \partial l_i} $ .\footnote{We omit the dependence of the loglikelihood on the hyperparameter $\theta$ for brevity of exposition here.} and the main diagonal entries of the Hessian $ \frac{\partial^2 \log Pr (y |l) }{ \partial l_i ^2} $ are provided in the Appendix section. Generally we have to compute $\frac{\partial K}{d l_i}$ and  $\frac{\partial^2 K}{d l^2_i}$ respectively. However, for each $i$ to  compute the gradient of the covariance matrix $K$ with respect to $l_i$ we have to compute $\frac{\partial K_{j,k}}{ \partial l_i} ; \forall j,k  \in \{ 1,2,..,N \}     \ st : j \geq i \geq k-1 $ since $K_{jk}$ is a function of $exp(j)+exp(j+1)+..+exp(k-1)$. This would take $O(N^2)$ for each $i$ and $ O(N^3) $ in total for all $i$. The same problem happens to calculation of the Hessian, when we need to figure out  $ \frac{\partial^{2} K_{j,k}}{ \partial l_i^2} ; \forall      j \geq i \geq k-1 $ .
	
	Nevertheless, we can use the intermediate results of $\frac{\partial logPr (y |l) }{ \partial l_{i-1}} $ to compute  $\frac{\partial \log Pr (y |l) }{ \partial l_i} $ thus reducing the total computation time  as follows. First, note that $  \ \frac{\partial K_{jk}}{ \partial \tau_h}=0  \   \textrm{for} \ h \notin \{j,k\}    \textrm{and}  \   \frac{\partial \tau_h}{ \partial l_i}=0  \    \textrm{for} \ h -1 \geq i   , \ \frac{\partial \tau_h}{ \partial l_i}=e^{l_i} \    \textrm{for} \ h  \leq i     $. Second according to the chain rule: 
	
	\begin{equation}  \frac{\partial K_{jk}}{ \partial l_i} = \sum_{h=1}^N  \frac{\partial K_{jk}}{ \partial \tau_h}    \frac{\partial \tau_h}{ \partial l_i} = \sum_{h=1}^i \frac{\partial K_{jk}}{ \partial \tau_h}    \frac{\partial \tau_h}{ \partial l_i} 
	=\big(   \sum_{h=1}^{i-1} \frac{\partial K_{jk}}{ \partial \tau_h}     \big)e^{l_i} +\frac{\partial K_{jk}}{ \partial \tau_i}  e^{l_i}  \end{equation}
	
	Thus, $ \frac{\partial K}{ \partial l_i}=  \frac{\partial K}{ \partial l_{i-1}} \frac{e^{l_i}}{e^{l_{i-1}}}+  \frac{\partial K}{ \partial \tau _i} e^{l_i} $ . Since computing of $\frac{\partial K}{ \partial \tau _i} $ takes $O(N)$ time so does $\frac{\partial K}{ \partial \l_i} $, overall  $\frac{\partial logPr (y |l) }{ \partial l} $ can be determined in $O(N^2)$ time.

	The same trick can be applied to compute $\frac{\partial^2 K}{\partial l_i^2}$, we have:

	\begin{equation}     \frac{\partial^2 K_{jk}}{ \partial l^2_i}= \sum_{h=1}^i \bigg( \frac{\partial K_{jk}}{ \partial \tau_h}    \frac{\partial \tau^2_h}{ \partial l^2_i} +\frac{\partial K^2_{jk}}{ \partial \tau^2_h} \big( \frac{\partial \tau_h}{ \partial l_i}  \big)^2   \bigg)+
	\sum_{  1 \leq h \neq h' \leq i} \bigg( \frac{\partial K^2_{jk}}{ \partial \tau_h \partial \tau_{h'}}  \frac{\partial \tau_h}{\partial l_i}\frac{\partial \tau_{h'}}{\partial l_i}  \bigg) \end{equation}

	In addition, we know that $K_{jk}$ can be considered as  a symmetric stationary kernel function of $\tau_j$ and $\tau_k$, $k(\tau_j; \tau_k)= k(\tau_j +\Delta; \tau_k+ \Delta) , \forall \Delta \in R$,   so $\frac{\partial^2 K_{jk}}{\partial \tau^{2}_j }+\frac{\partial^2 K_{jk}}{\partial \tau^{2}_k} +\frac{\partial^2 K_{jk}}{\partial \tau_k \partial \tau_j }+\frac{\partial^2 K_{jk}}{\partial \tau_j \partial \tau_k }=0$ . Based on this observation to compute $\frac{\partial^2 K}{ \partial l^2_i} $we only need to compute  $ \frac{\partial^2 K_{h i}}{\partial \tau^2_i} $ for $h>i$ at i-step and it costs $O(N)$ time. Hence the overall complexity for calculating $ \frac{\partial^2 K}{ \partial l^2_i}; \forall i$ is $O(N^2)$.

	\section{Baseline method}
	In order to demonstrate the benefits of our proposed model, we will compare our model with MCMC, NIGP and GP in experiments. The detail of MCMC-based method for handling noisy inputs with ordering constraints is given below.
	
	\subsection{Monte Carlo Markov Chain sampling}
	
	We can use MCMC for drawing samples of $\tau$ and $\theta$ from the joint posterior distribution $Pr(\tau; \theta | y, t, \sigma_t) \propto Pr(y|\tau; \theta) Pr(t | \tau, \sigma_t) Pr(\tau) Pr(\theta)$. For simplicity, we assume both $\tau$ and $\theta$ have a uniform prior distribution. 
	
	We follow Metropolis-Hasting sampling stragey here, when a Gaussian distribution will be used as  the proposal distribution. In particular, at (k+1)-th iteration for each $i \in \{1,2,..,n\}$ we generate a new potential true inputs $\tau_i^{k+1}$ using a Gaussian function $q(\tau_i^{k+1}; \tau_i^{k})$ centered at 
	$\tau_{i}^{k} $. Since we have ordering constraints over the true inputs, i.e $\tau_{i-1}>\tau_{i}> \tau_{i+1} $ so  we propose to discard this sample if it violate the constraints and continue to the next index of inputs, $i+1$. Otherwise, we will accept this sample with the probability
	\begin{equation} min \bigg(1, \frac{Pr \big( \tau_{i}^{k+1}| y, t, \theta^{k},    S \setminus \{\tau_i^{k}\} \big) q(\tau_i^{k};\tau_i^{k+1}  \big)       }{ Pr \big( \tau_{i}^{k}| y, t, \theta^{k},    S \setminus \{\tau_i^{k}\} \big) q(\tau_i^{k+1};\tau_i^{k}  )} \bigg) \end{equation}
	
	Where $S$ denotes the set of currently stored samples for $\tau$, $card(S)=N$. 
	Note that in order to compute the acceptance probability, we need to compute the likelihood of $y$ given the currently stored samples without one element, and this costs $O(N^3)$ operations for inverting the covariance matrix. We can reduce the time complexity at each step as follows. Note that when we iterate each location $i$ of $\tau$ in turn, the covariance matrix will  change only on i-th row and  i-th column.   Based on this observation we can reduce the running time of MCMC by  applying the following Woodbury matrix inversion lemma: 
	\begin{equation} (K+uv^T)^{-1}= K^{-1}- \frac{K^{-1} u v^T K^{-1}}{1+v^T K^{-1} u}\end{equation}
	
	This reduces the complexity of updates from $O(N^3) $ to $O(N^2)$.

	\section{Experiments}
	\label{exp}
	In this section we first demonstrate the effectiveness of our proposed method on artificial datasets since there is no available real datasets with groundtruth. Then we run the proposed model with MCMC, NIGP and GP to see if there is any difference on Northeastern Florida dataset where no groundtruth exists.
	\subsection{ Evaluation on synthetic datasets}
	
	Here we consider the set of experiments that were suggested in \cite{NIPS2011_4295} with the added ordering constraints on the latent input variables $\tau$. In particular, there are 25 input points  that are equally spaced in $[-10;10]$ and we are trying to learn functions that have varying gradients across the input space. The range of all functions are the same, 10. The reason for doing this is we want to see behavior of methods in two cases: small and large output noise. For the output noise, we consider a small $\sigma_y = 0.05$ and a large $\sigma_y = 1$ noise setting.  The input noise will vary in the range $\sigma_t \in [0.2, 3]$.
	\begin{figure}[t] 
		\captionsetup[subfigure]{aboveskip=-1pt,belowskip=-1pt}
		\begin{subfigure}[b]{0.33\linewidth}
			\centering
			\includegraphics[width=0.95\linewidth]{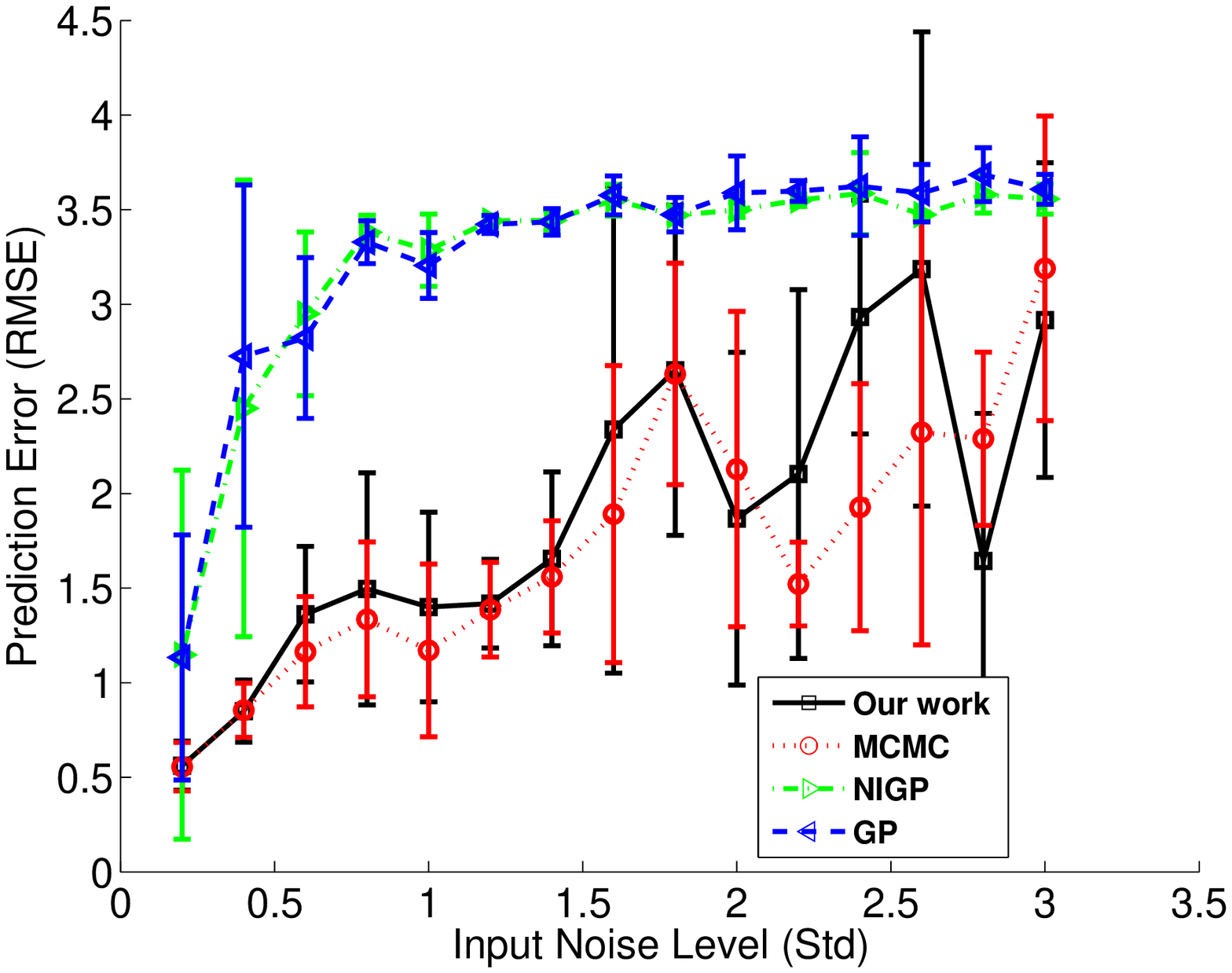} 
			\caption{ $ 5 \sin(\tau) $} 
			\label{fig7:a} 
			\vspace{2ex}
		\end{subfigure}
		\begin{subfigure}[b]{0.33\linewidth}
			\centering
			\includegraphics[width=0.95\linewidth]{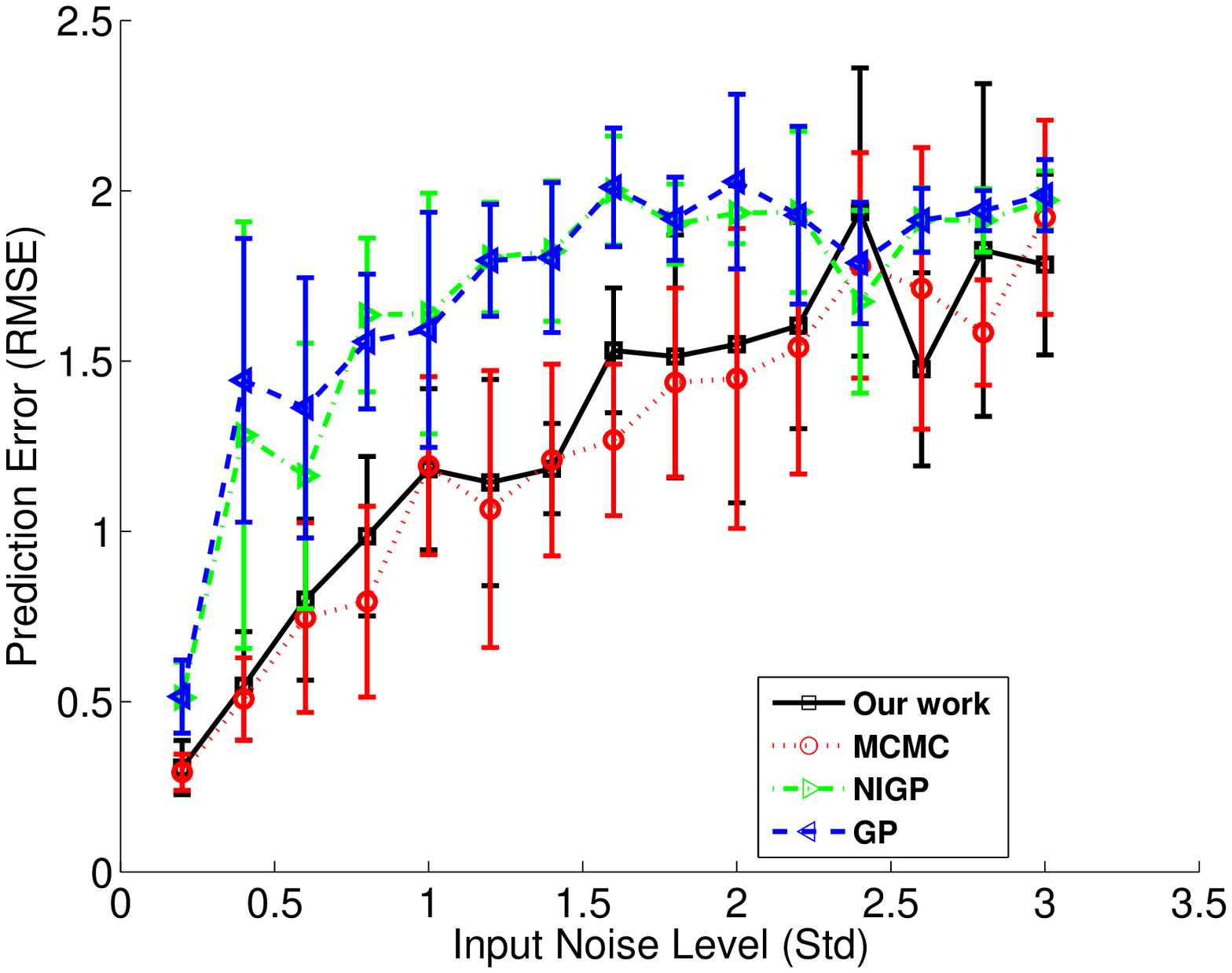} 
			\caption{$ 1.147 e^{-0.2 \tau} \sin(\tau)$} 
			\label{fig7:b} 
			\vspace{2ex}
		\end{subfigure} 
		\begin{subfigure}[b]{0.33\linewidth}
			\centering
			\includegraphics[width=0.95\linewidth]{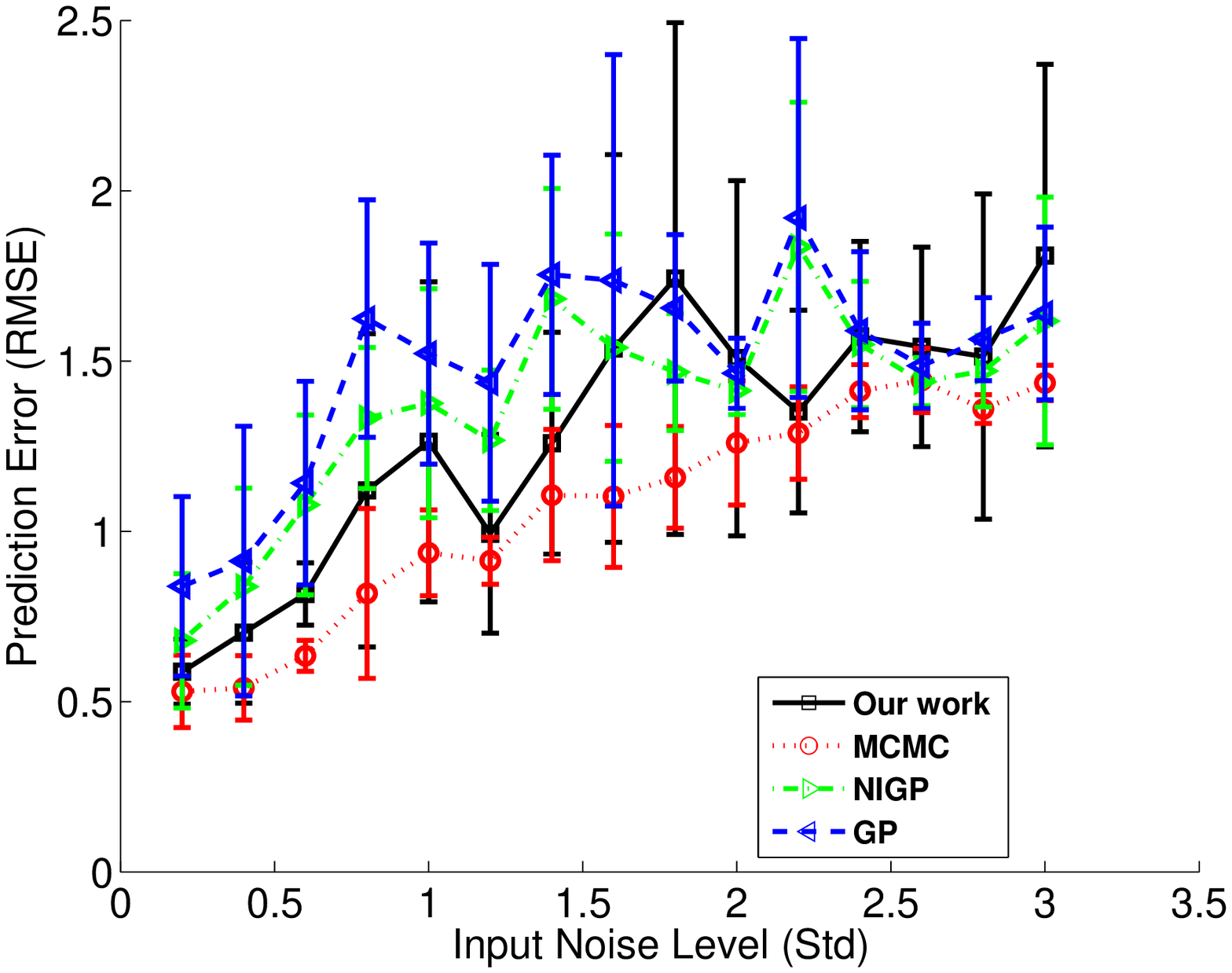} 
			\caption{$ 0.97 \tan(0.15 \tau) \sin(\tau)$} 
			\label{fig7:c} 
			\vspace{2ex}
		\end{subfigure}
		
		\begin{subfigure}[b]{0.33\linewidth}
			\centering
			\includegraphics[width=0.95\linewidth]{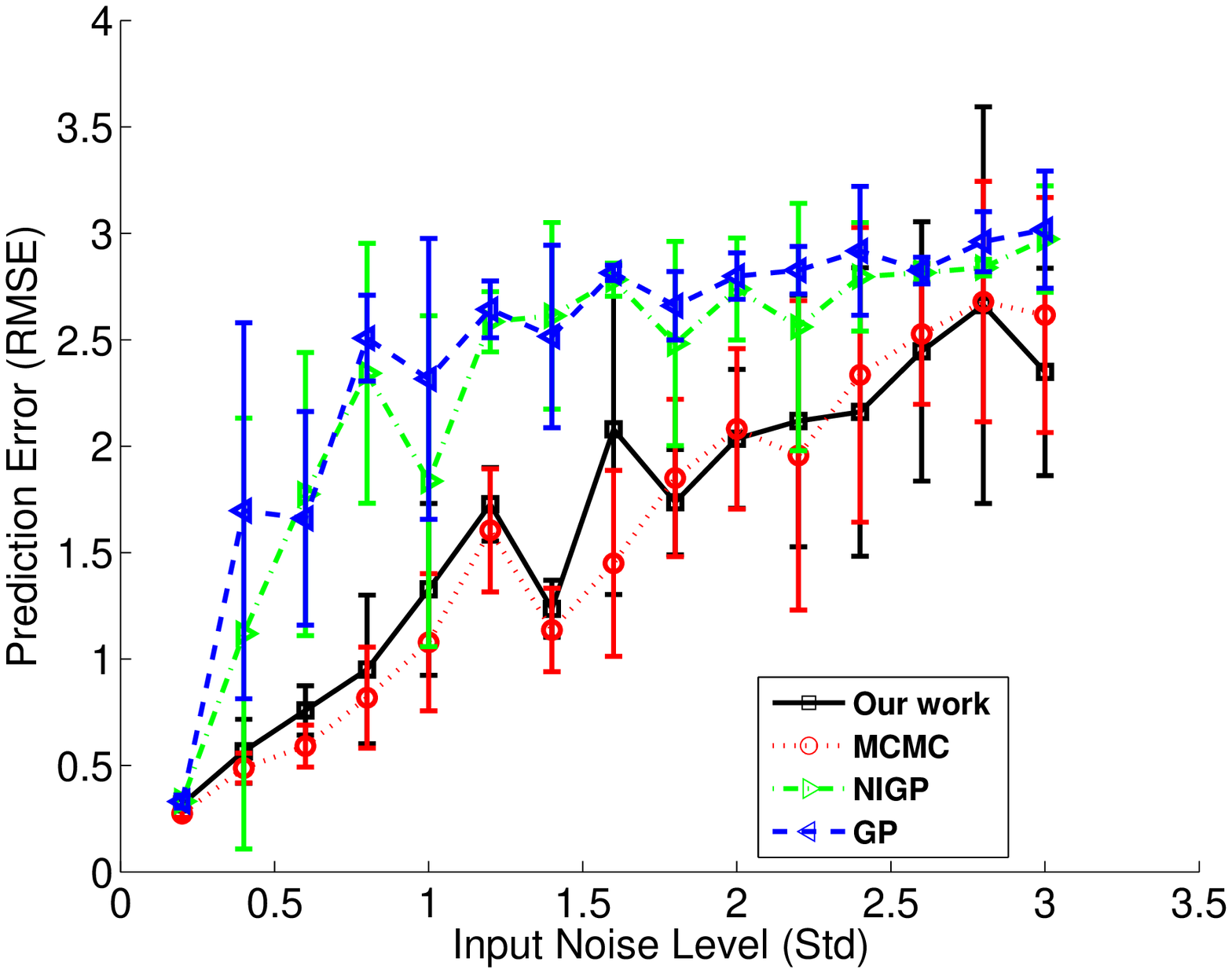} 
			\caption{$0.055 \tau^2 \tanh (cos(\tau))$ } 
			\label{fig7:c} 
			\vspace{2ex}
		\end{subfigure}
		\begin{subfigure}[b]{0.33\linewidth}
			\centering
			\includegraphics[width=0.95\linewidth]{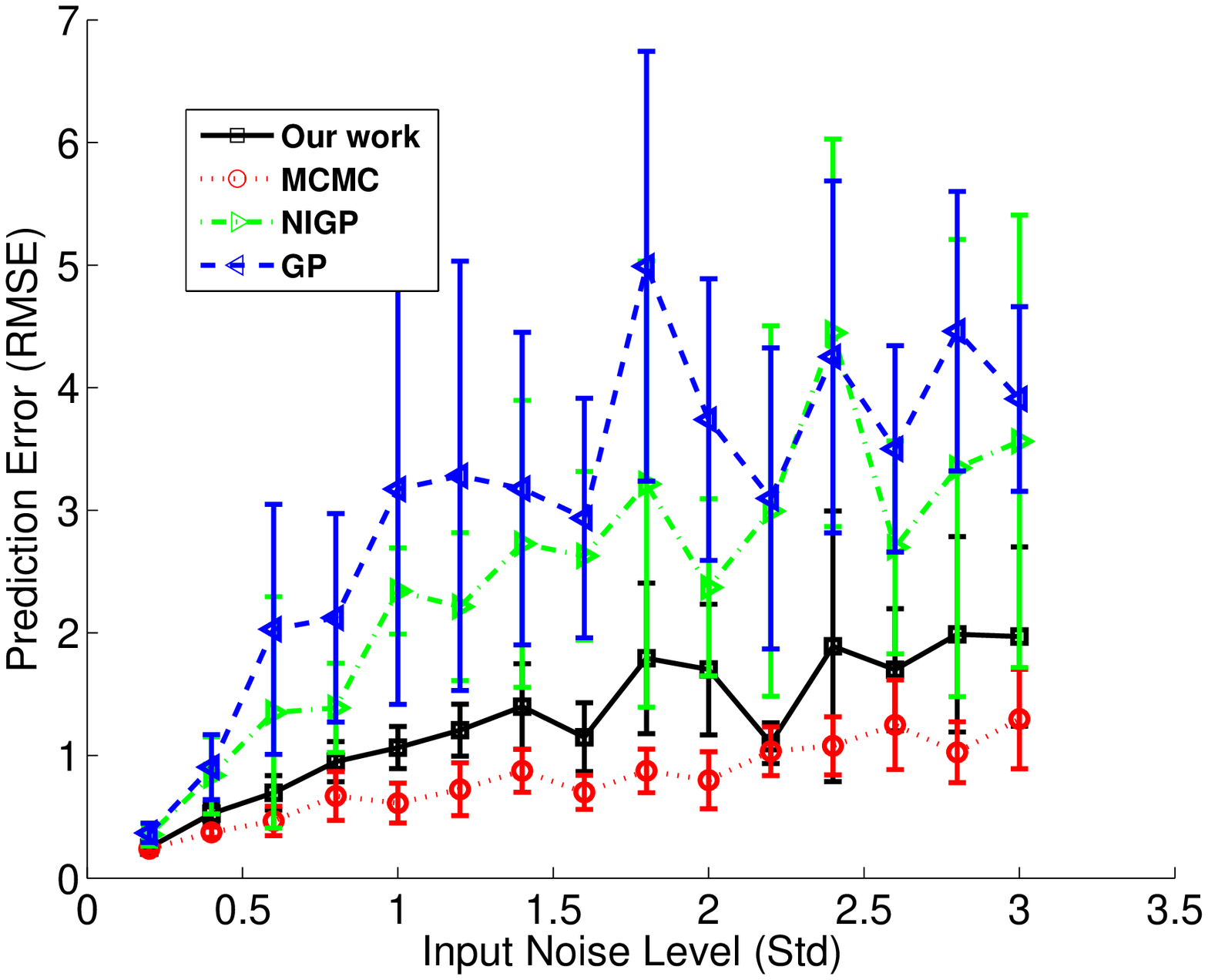} 
			\caption{$1.76 \log( \tau^2 (\sin (2\tau)+1)+1)$} 
			\label{fig7:a} 
			\vspace{2ex}
		\end{subfigure}
		
		\begin{subfigure}[b]{0.33\linewidth}
			\centering
			\includegraphics[width=0.95\linewidth]{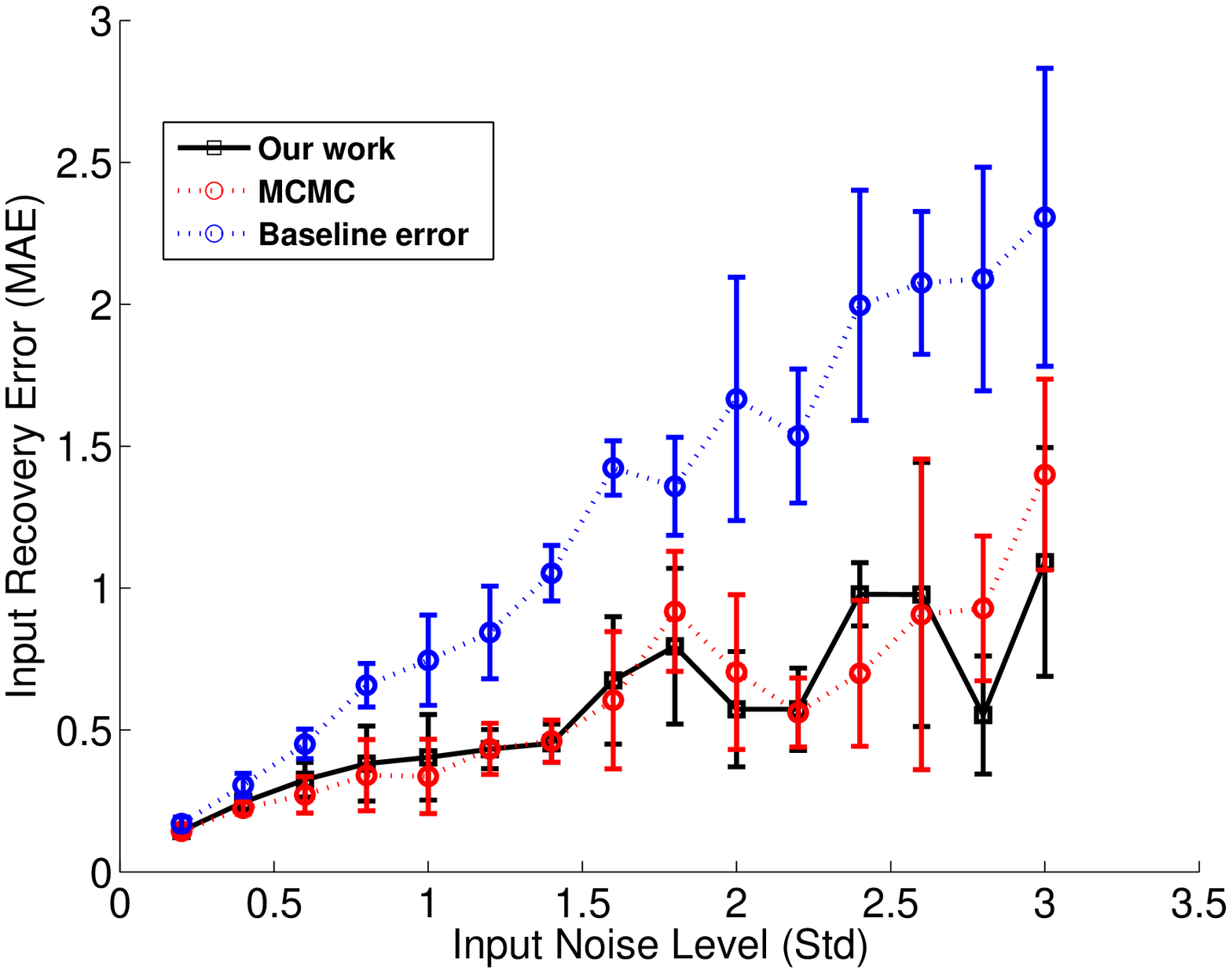} 
			\caption{$5 \sin(\tau)$} 
			\label{fig7:b} 
			\vspace{2ex}
		\end{subfigure}
		\begin{subfigure}[b]{0.33\linewidth}
			\centering
			\includegraphics[width=0.95\linewidth]{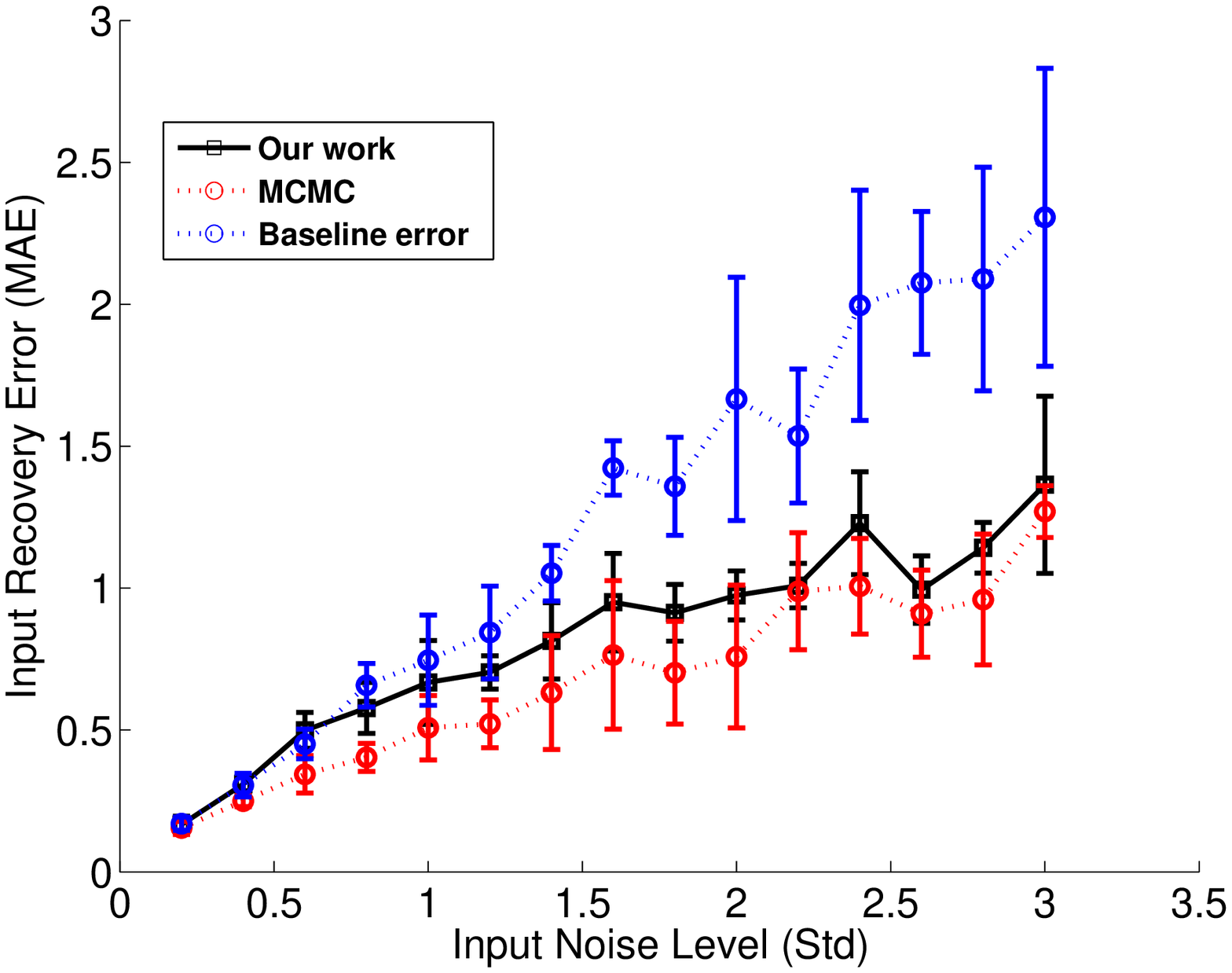} 
			\caption{$ 1.147 e^{-0.2 \tau}sin(\tau)$} 
			\label{fig7:c} 
			\vspace{2ex}
		\end{subfigure}
		\begin{subfigure}[b]{0.33\linewidth}
			\centering
			\includegraphics[width=0.85\linewidth]{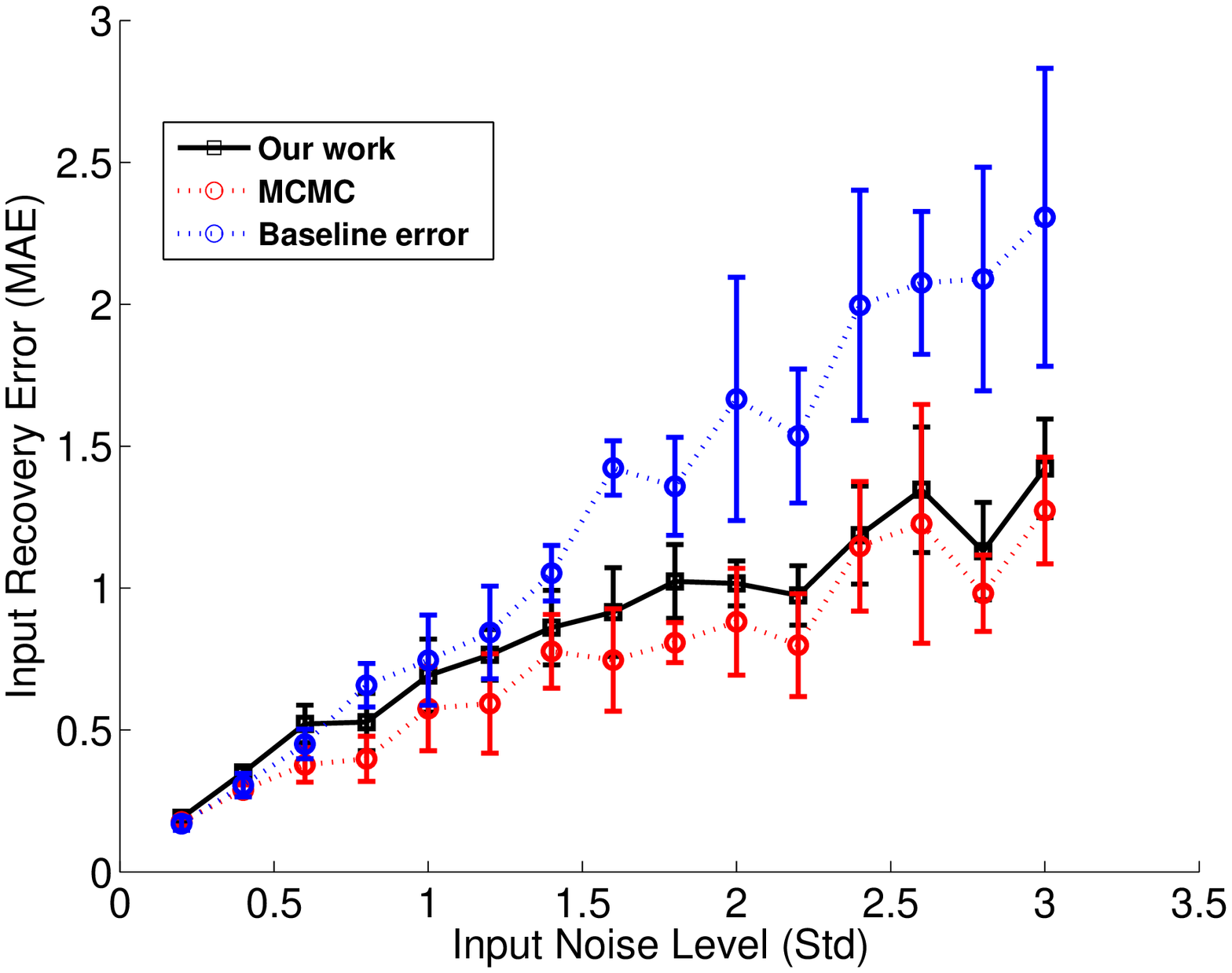} 
			\caption{$0.97 \tan(0.15 \tau) \sin(\tau)$} 
			\label{fig7:c} 
			\vspace{2ex}
		\end{subfigure}
		
		\begin{subfigure}[b]{0.33\linewidth}
			\centering
			\includegraphics[width=0.95\linewidth]{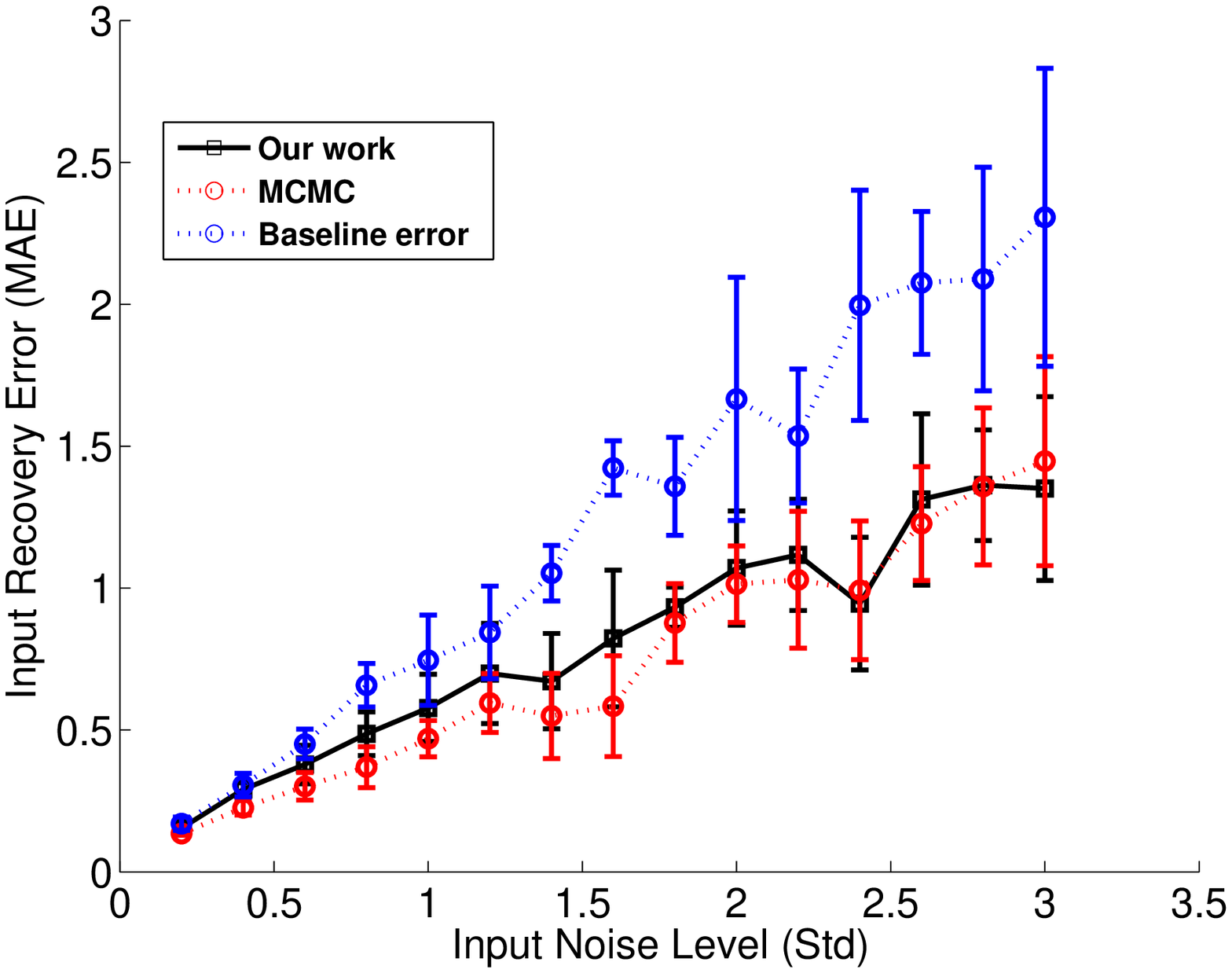} 
			\caption{$0.055 \tau^2 tanh (cos(\tau))$} 
			\label{fig7:c} 
			\vspace{2ex}
		\end{subfigure}
		\begin{subfigure}[b]{0.33\linewidth}
			\centering
			\includegraphics[width=0.95\linewidth]{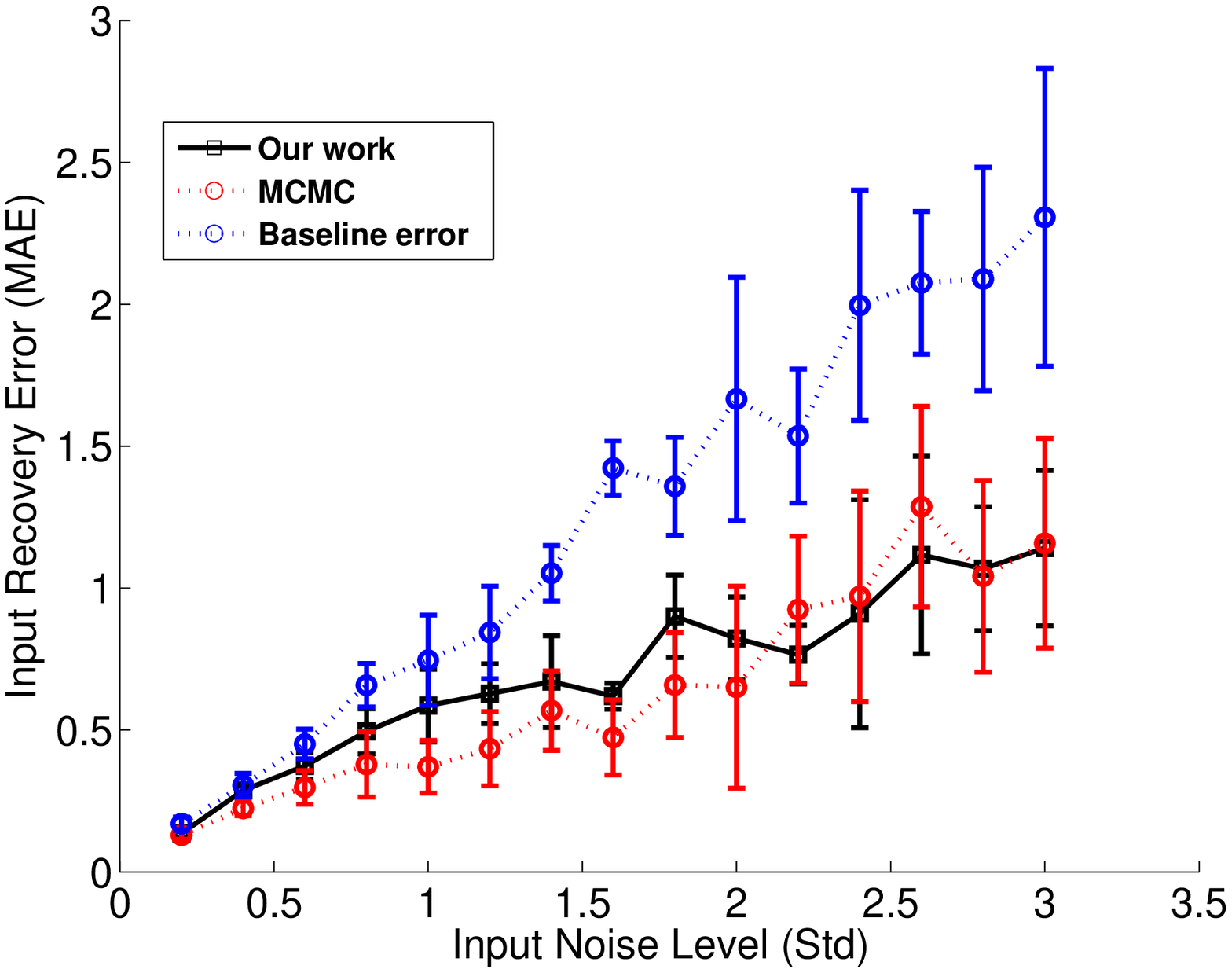} 
			\caption{$1.76 \log( \tau^2 (\sin (2\tau)+1)+1)$} 
			\label{fig7:c} 
			\vspace{2ex}
		\end{subfigure}
		
		\caption{ (a)-(e) Comparison among four methods: our proposed approach, MCMC, NIGP and GP on five latent functions based on prediction error. (f)- (j) Comparison between our work and MCMC in term of the ability to estimate the true inputs. The output noise is a small constant value: $\sigma_y=0.05$, while the input noise changes from $0.2$ to  $3$. The baseline error is the mean absolute difference between noisy inputs $t$ and $\tau$.}
		\label{fig7} 
	\end{figure}

	We employ a Matern covariance with here because it was used in previous works, eg [8] for sea level modelling 
	$$ C(\tau_i; \tau_j)_{\nu=\frac{3}{2}}= \sigma^2_f \big(  1 + \frac{\sqrt{3} | \tau_i- \tau_j|}{d}\big) e^{- \frac{-\sqrt{3} |\tau_i -\tau_j |}{d}} ; \  \theta={\sigma_f; d} $$
	We will compare four methods: our proposed method, MCMC, NIGP and usual GP. Experiments settings for each method are as follows:
	
	For our proposed model we used a mixture of Gaussians with $K=3$ for the variational distribution $Q(l,r)$ and scaled conjugate gradient to minimize $F(\Phi)$ in Eq.(12). To avoid local minima, we run our model five times each time with different inital points of parameters and choose the model that return the smallest objective function value. We set a default value of 5000 iterations for MCMC since based on our experiments, this value is large enough for convergence to the stationary distribution. For NIGP we used the Matlab's global optimization toolbox to learn the GP hyperparameters.  The average training time (s) after five single runs for our proposed method, MCMC, NIGP and GP alternatively are $19 (\pm 1.2), 61( \pm 5.5), 106 (\pm 4.6), 0.3 (\pm 0.008)$. This confirms that our proposed method is computationally significantly more efficient than both NIGP and MCMC.
	
	Next, we compare four methods based on prediction errors in the case of small output noise. The criateria for comparison is the root mean squared error (RMSE). For each case of input noise level, we run each method five times, then take the average prediction errors with the standard deviations. Fig (1). (a)-(e) indicates the peformance of four methods for each function.
	
	Based on those figures, Our proposed method demonstrates predictive performance on par with MCMC, while being significantly more computationally efficient.  The method consistently outperforms GP and NIGP that do not utilize the ordering constraints, and hence are unable to effectively deal with the input noise.   To test the ability of different methods to recover the true input, we examined the input estimation errors for MCMC and our proposed approach. This is depicted in Fig.(1) f-j.  Note that neither GP nor NIGP explicitly seek to recover the input estimates.  As the evaluation criteria, we use the mean absolute error(MAE) and contrast it with the amount of noise in the input.  Experimental results indicate that both MCMC and our proposed approach effectively reduce the amount of noise in the input, which, in turn, enables more accurate function prediction.  Again, our proposed approach accomplishes this task in a computationally more efficient manner than the competing MCMC
	
	However, when the output noise is large, $\sigma_y=1$, the improvement of our method over NIGP and GP is not much. Only three out of five cases of selected functions, our work performs clearly better than GP with different input noise levels. In the last case of function $f(\tau)=1.76 \log(\tau^2 \sin(2\tau+2)+1)$, NIGP even has smaller prediction errors than our work. We provide experimental results of this case in supplementary material. 
	
	\subsection{Application to sea level estimation}
	
	We demonstrate the application of the proposed model on the reconstruction of sea level in Notheastern Florida  from 700 BC to 2010 AD. We used the dataset that was provided in supplenemtary data described in \cite{kemp2014late}, consisting of 77 data points, ranging from 560 BC to 2010 AD. Among them 65 instances have noisy inputs. The standard deviation of output noise at the 65 noisy input measurements  is constant and very large, $\sigma_{y}=181 \approx \frac{max(y)-min(y)}{10}$. The output noise at the remaining 12 instances is smaller. We used the same settings of  all methods as the previous experiments with synthetic datsets.
	
	Prediction results for the four methods are displayed in Fig. (2).  A more insightful look can be gained by considering at the differences in the predictions of the four methods. In Tab. (1) we show the average absolute pairwise differences between mean predictions of different approaches, together with the average symmetrized KL divergence of predictive densities on query points.  All differences are statistically significant at 5\% level, indicating that different methods make could lead to alternative explanations of the sea-level rise history as well as result in different predictive models.
	
	\begin{table}[H]
		\captionsetup{justification=centering,margin=1cm}
		\caption{Mean absolute error and symmetrized KL divergence for measuring the difference between predictions of four models}
		\begin{minipage}[t]{.48\linewidth}
			\centering 
			\subcaption{Mean absolute differences between mean predictions of any pair of four methods}
			\begin{tabular}{lllll}
				\multicolumn{1}{c}{\bf } &\multicolumn{1}{c}{\bf NVP}  &\multicolumn{1}{c}{\bf MCMC} &\multicolumn{1}{c}{\bf NIGP}  &\multicolumn{1}{c}{\bf GP}
				\\ \hline \\
				\bf NVP        & - & 1.23  &3.49  & 3.42\\   
				\bf MCMC      &   - &-   & 2.80 & 2.73\\
				\bf NIGP         & - & - &-  & 0.069\\
				\bf GP            &- & - & - & - \\
			\end{tabular}
		\end{minipage}
		\hfill
		\begin{minipage}[t]{.48\linewidth}
			\centering
			\subcaption{Symmetrized KL divergence between predictive posterior distribution of any pair from four methods} 
			\begin{tabular}{lllll}
				\multicolumn{1}{c}{\bf } &\multicolumn{1}{c}{\bf NVP}  &\multicolumn{1}{c}{\bf MCMC} &\multicolumn{1}{c}{\bf NIGP}  &\multicolumn{1}{c}{\bf GP}
				\\ \hline \\
				\bf NVP        & - & 0.71  & 3.2 & 3.02\\   
				\bf MCMC      &   - & -  & 0.98 & 0.88\\
				\bf NIGP         & - & -  & -  & 0.003\\
				\bf GP            &- & - & - &- \\
			\end{tabular}
		\end{minipage} 
	\end{table}
	Our model could be used to predict the sea level rise rate in the future. These outcomes are of particular concerns in the context of climate science research and suggesting possible reactions to the threat of the sea level rise. However, this should be done with climate-driven data which we do not have for now and we will leave this for our future work.

	\begin{figure}[H] 
		\begin{subfigure}[b]{0.5\linewidth}
			\centering
			\includegraphics[width=0.85\linewidth]{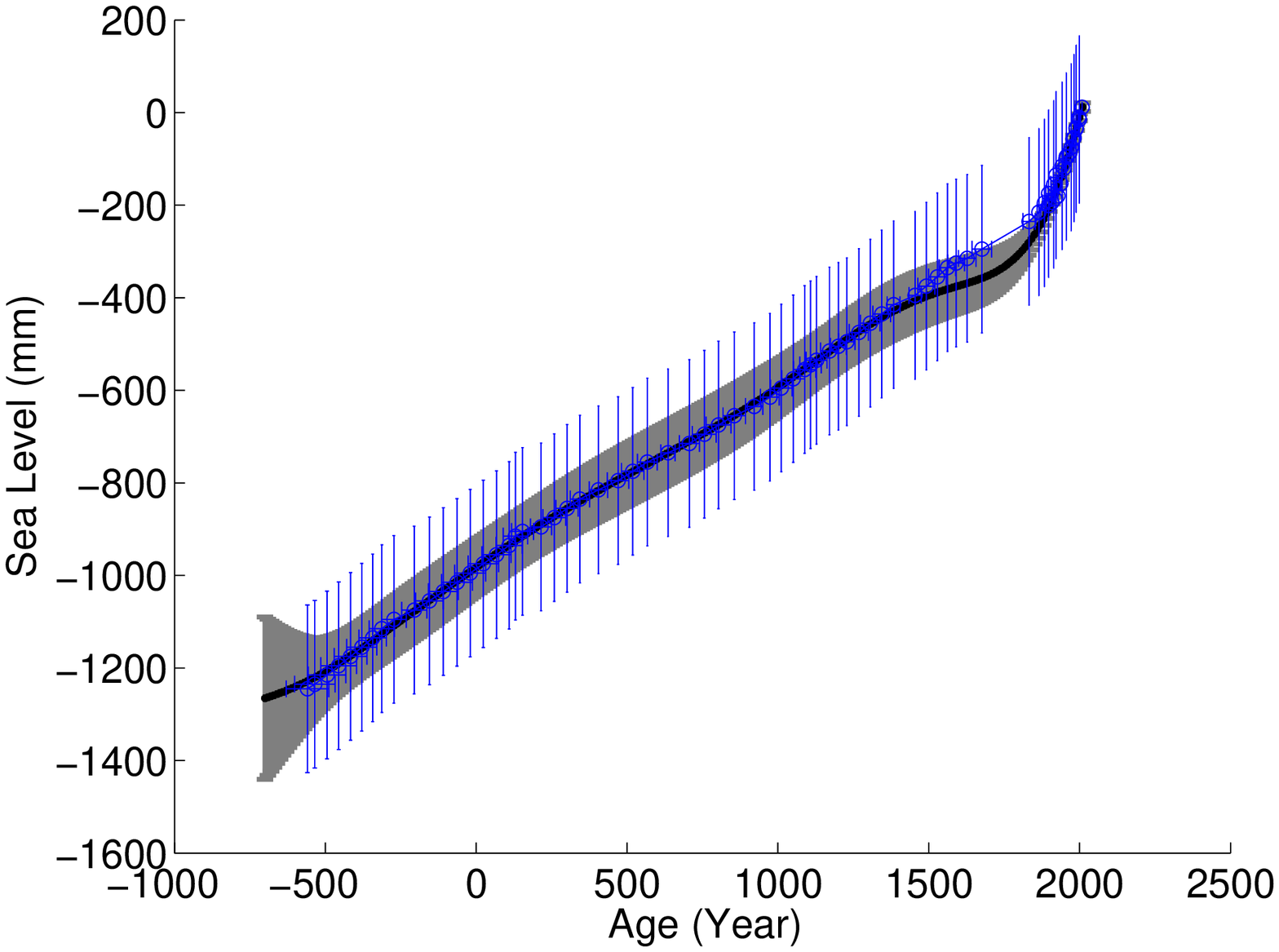} 
			\caption{Sea level reconstruction by NPV} 
			\label{fig7:a} 
			\vspace{2ex}
		\end{subfigure}
		\begin{subfigure}[b]{0.5\linewidth}
			\centering
			\includegraphics[width=0.85\linewidth]{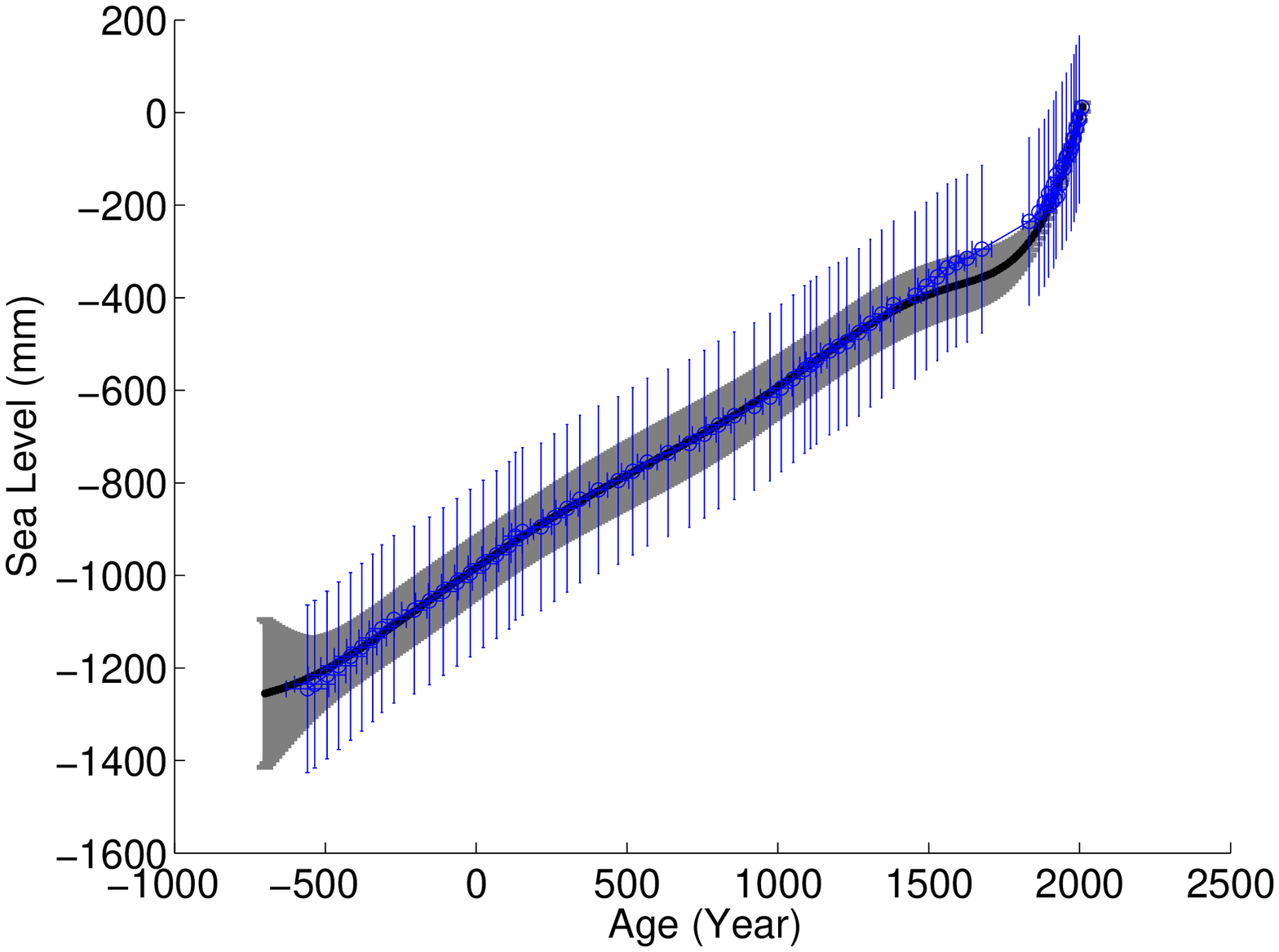} 
			\caption{Sea level rise reconstruction by MCMC} 
			\label{fig7:b} 
			\vspace{2ex}
		\end{subfigure}
		
		\begin{subfigure}[b]{0.5\linewidth}
			\centering
			\includegraphics[width=0.85\linewidth]{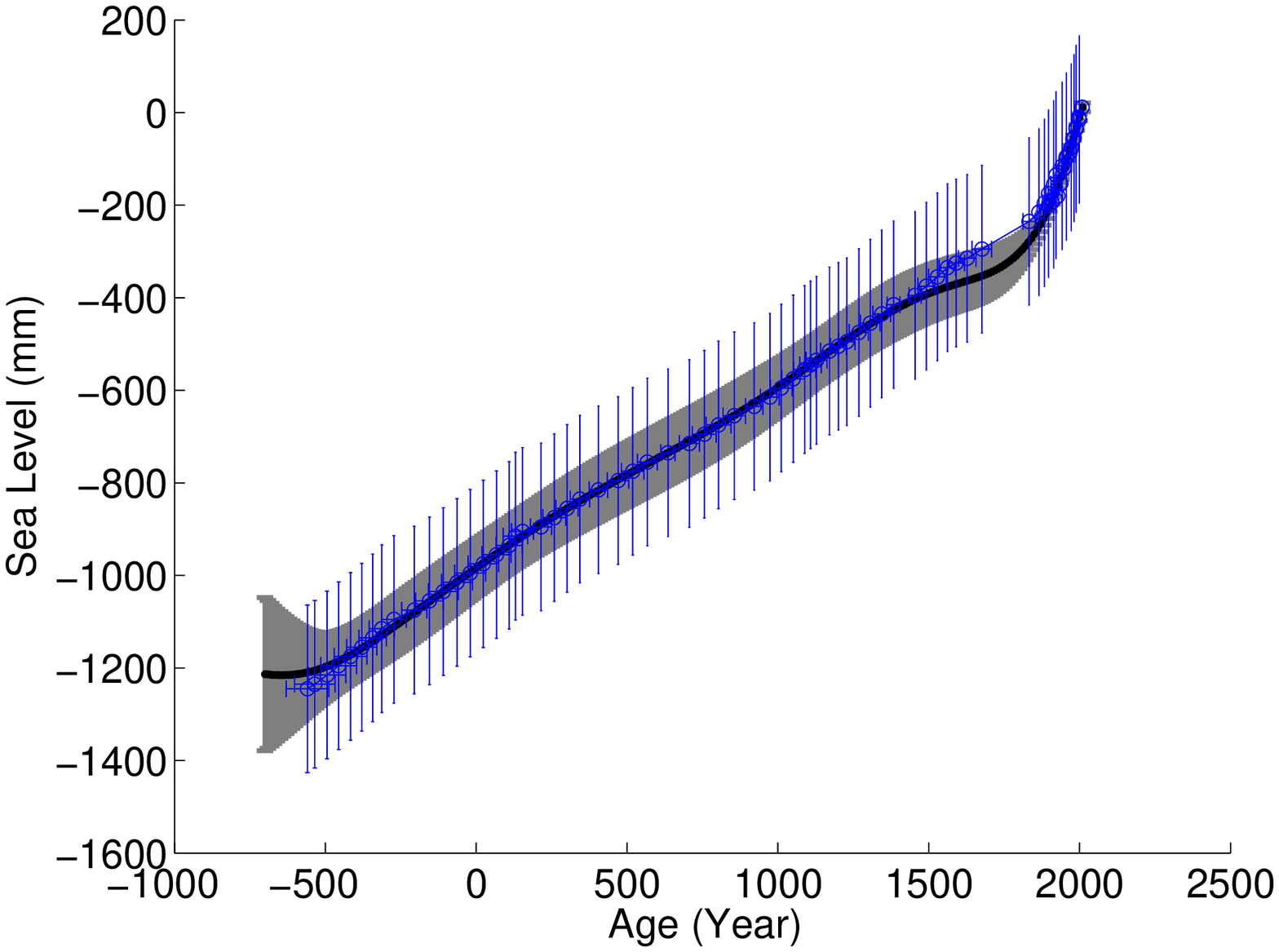} 
			\caption{Sea level reconstruction by NIGP} 
			\label{fig7:c} 
			\vspace{2ex}
		\end{subfigure}
		\begin{subfigure}[b]{0.5\linewidth}
			\centering
			\includegraphics[width=0.85\linewidth]{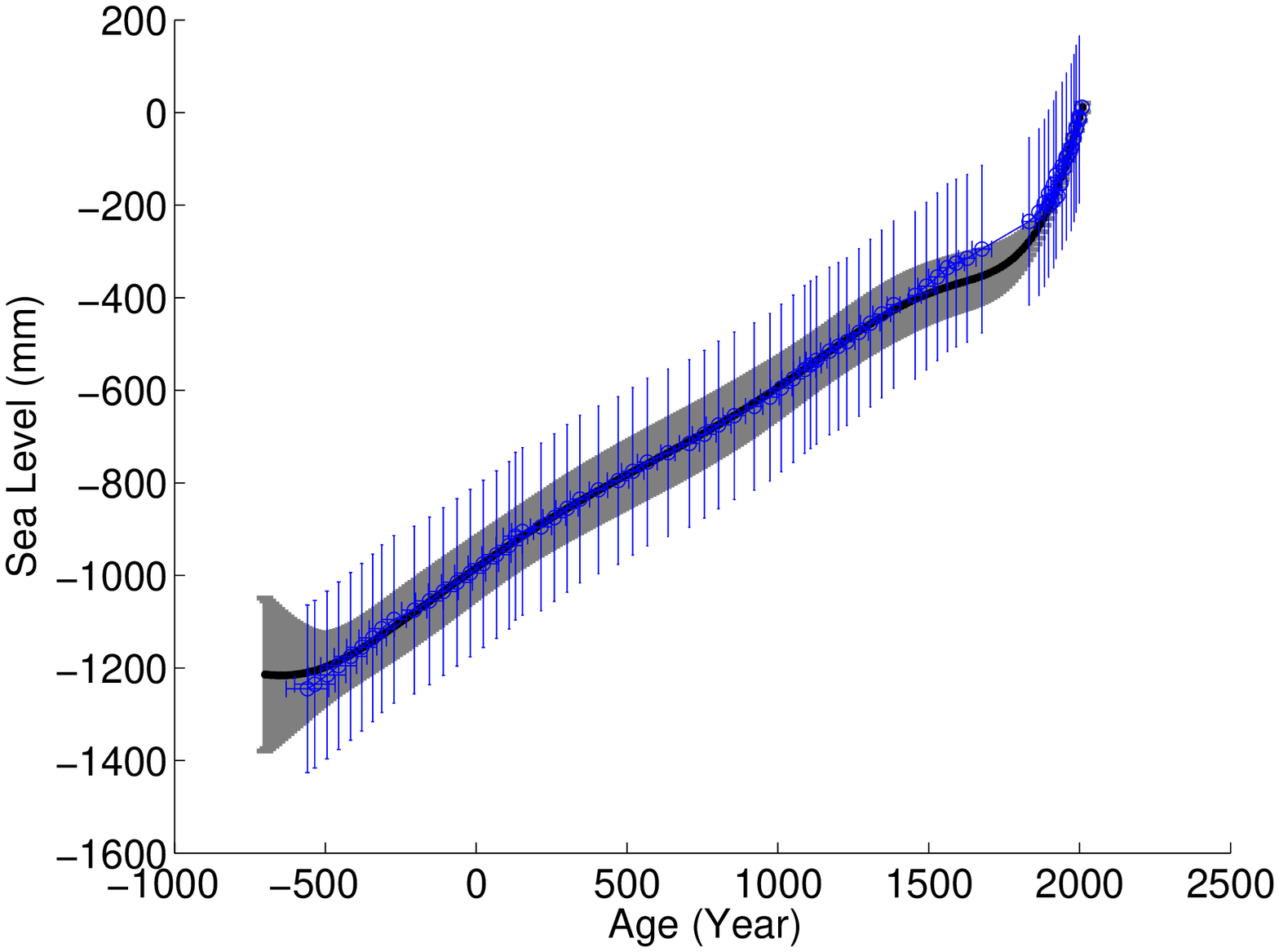} 
			\caption{Sea level reconstruction by standard GP} 
			\label{fig7:d} 
			\vspace{2ex}
		\end{subfigure} 
		\caption{Reconstruction of Northeastern Florida sea level using four methods. The noisy observations $(t,y)$ with input and output noise level were marked by a blue curve and the error bars in both directions indicate the input and output noise level. The mean predictions with $\pm$ one standard deviation for each method were also plotted. It seems that there is little diffence among methods.. }
		\label{fig7} 
	\end{figure}

	\section{Conclusions and discussion}
	
	In this paper, we have introduced an efficient and effective GP-based method to handle noisy inputs when we knew the order of the latent true inputs variables. We transformed those latent variables to obtain unconstrained ones and a Bayesian treatment was applied to infer the posterior distribution of the new variables.  The experiments indicated the improvement of our model over NIGP in term of running  time and prediction error.   Our model assumes that the input and output noise level are known because these quanties are given in sea level domains. In other practices, however we can consider them as parameters and modify the objective function to optimize them.  Our future research work involves applying the proposed model to reconstruct global sea level. In this case, besides temporal inputs we have spatial information, the  location of sea level, the problem becomes more complicated.
	
	{\small
		\bibliographystyle{unsrt}
		\bibliography{cuong}
	}
	

\end{document}


\maketitle

\section{An exact closed form of $ E_{Q}  log Pr(t | l,r, \sigma_t ) $ }
The variational distribution $Q$ is a mixture of $K$ Gaussians:  $Q=\frac{1}{K} \sum_{k=1}^K Q_i  ; Q_{i}= \mathcal{N}(l,r |m_{i}, V_{i})$ and  $E_{Q_i} [l_j] =m_{i,j} ;   E_{Q_i} [r] = m_{i,n} ; \forall i=1,2,..,K  ; \  \forall j=1,2,..n-1 $

 It is sufficent to provide a closed form for $E_{Q_k}log Pr(t | l,r, \sigma_t ) $ for any $k$ here. 
 First, for any Gaussian random variable $x \sim Pr(x)=\mathcal{N}(m_x,v_x)$ we have  $E_{Pr(x)}e^x =e^{m_x +\frac{v_x}{2}}$. Let $  \langle .  \rangle $  be the expectation operator with respect to the distribution $Q_k$. Define the following terms:

\begin{equation}   L_i = \langle  \sum_{j=i}^{i} l_j  \rangle= \sum_{j=i}^{i} e^{m_{k,j} + \frac{v_{k,j}}{2}} \         \textnormal{and}  \  S_i= \sum_{j=1}^i\frac{1}{\sigma_{t,j}^2}   ; \forall i=1,2,..,n\end{equation}

It is easy to verify that: 

\begin{equation} 
\begin{aligned}
 \left \langle  log Pr(t |l,r, \sigma_t)   \right \rangle  =\left \langle log \prod_{i=1}^{n}   \bigg( \mathcal{N}(t_i | r+ \sum_{i=0}^{n-i}  l_i; \sigma_{t,i})  \bigg)  \right \rangle  &= \left\langle   \sum_{i=1}^{n-1} - \frac{  \big( r + \sum_{j=i}^{i}e^{l_j} -t_i \big)^2 }{2 \sigma^2_{t,i}}  \right \rangle - \left  \langle \frac{ \big( r-t_n \big)^2}{2\sigma^2_{t,n}} \right \rangle 
\\=P1+P2+P3+P4 +const
\end{aligned}
\end{equation}

Where:

\begin{equation}  P_1=-\frac{1}{2} \sum_{i=1}^n \frac{\langle r^2  \rangle + 2 \langle r  \rangle \langle  \sum_{j=1}^{i} e^{l_j}   \rangle  }{\sigma_{t,i}^2} = -\frac{1}{2} \bigg(  S_n( m_{k,n}^2+v_{k,n}) +2m_{k,n} \sum_{i=1}^n \frac{L_{i}-t_{i}}{\sigma_{t,i}^2} \bigg)    \end{equation}
\begin{equation}  P_2=-\frac{1}{2} \sum_{i=1}^n  \left \langle  \frac{ \sum_{j=1}^i (e^{l_j})^2 }{\sigma_{t,i^2}}   \right \rangle =-\frac{1}{2} \bigg(  \sum_{i=1}^{n-1} e^{v_{k,i}} (   e^{2m_{k,i} +v_{k,i}}   ) S_i \bigg)  \end{equation}
\begin{equation} P_3 =-\frac{1}{2} \left \langle \frac{\sum_{ 1 \leq j  \neq j' \leq i }2 e^{l_{j}} e^{l_{j'}}    }{\sigma_{t,i}^2} \right \rangle =-\sum_{i=1}^{n-2} \sum_{j=i+1}^{n-1}   e^{\frac{v_{k,i}+v_{k,j}}{2} +m_{k,i}+m_{k,j}} S_{i}    \end{equation}
\begin{equation}  P_4=-\frac{1}{2} \sum_{i=1}^n  \left \langle  \frac{2 t_i \big(  \sum_{j=1}^i e^{l_j} \big) }{\sigma_{t,i^2}}   \right \rangle =\sum_{i=1}^n \frac{t_i L_i}{\sigma_{t,i}^2}  \end{equation}

\section{Gradient and Hessian of $log Pr (y |l , \theta)$ with respect to $l$}
First, let $\gamma= K^{-1}y$ and $T= \gamma \gamma^T -K^{-1}$ 
The gradient of $log Pr (y | l, \theta) = -\frac{1}{2}y^T K y  -log | K | + const$ with respect to $l_i$ is:

\begin{equation}\frac{\partial log Pr( y |l , \theta)}{ \partial l_i} = \frac{1}{2} trace \bigg[  \big( \gamma \gamma^T -K^{-1} \big) \frac{\partial K}{\partial l_i}  \bigg]  \end{equation}

Define $M_i= \big( \gamma \gamma^T -K^{-1} \big) \frac{\partial K}{\partial l_i}=T \frac{\partial K}{\partial l_i}$ then  $ \frac{ \partial^2 log Pr(y |l , \theta) }{ \partial l_i^2} = \frac{1}{2} trace ( \frac{ \partial M_i}{\partial l_i})   $. We have: 

\begin{equation} \frac{\partial M_i}{\partial l_i} = \frac{\partial \big(  T \frac{\partial K }{ \partial l_i} \big) }{\partial l_i} =T \frac{\partial^2 K}{\partial l_i^2 } +\frac{\partial K}{\partial l_i} \frac{\partial T}{\partial  l_i}  \end{equation}

In addition, put $ I_i = \frac{\partial K^{-1}}{l_i}= -K^{-1} \frac{\partial K}{\partial l_i} K^{-1} $  and $Z_i =\gamma ( \frac{\partial \gamma}{\partial l_i})^T=\gamma (I_i y)^T $ .We can express the gradient of $T$ w.r.t $l_i$ by : 

 \begin{equation} \frac{\partial T}{\partial l_i} = \frac{\partial \big(   \gamma \gamma^T - K^{-1} \big) }{\partial l_i} =\frac{\partial \gamma}{l_i} \gamma^T +\big( \frac{\partial \gamma}{l_i} \big)^T \gamma -\frac{\partial K^{-1}}{\partial l_i}= Z_i+ Z_i^T + K^{-1} \frac{\partial K}{\partial l_i} K^{-1} \end{equation}

In Eq. (3), we applied the following formula: $\frac{\partial Y^{-1}}{x} =- Y^{-1} \frac{\partial Y}{\partial x} Y^{-1} $

\begin{equation}  \end{equation}

Overall, the procedures to compute  $ \frac{ \partial^2 log Pr(y |l , \theta) }{ \partial l_i^2}$  can be listed as below:

\begin{enumerate}
\item  Calculate  $\frac{\partial K}{\partial l_i} $  and $ \frac{\partial^2 K}{\partial \l_i^2} $ 
\item Calculate $  I_i = \frac{\partial K^{-1}}{\l_j}= -K^{-1} \frac{\partial K}{\partial \l_j} K^{-1}   $
\item  Calculate $ Z_i= \gamma (I_i y)^T$ 
\item Calculate $ \frac{ \partial M_i}{\partial l_i}= T \frac{\partial^2 K}{\partial l_i^2} +\frac{\partial K}{\partial l_i} ( Z_i+ Z_i^T -I_i) $
\item Output: $ \frac{ \partial^2 log Pr(y |l , \theta) }{ \partial l_i^2} = \frac{1}{2} trace ( \frac{ \partial M_i}{\partial l_i})   $ 
\end{enumerate}

\section{Additional experiments}

Here we demonstrate our proposed method still performs well  when the output noise is large, $\sigma_y=\frac{max(f)-min(f)}{10}$. We use the same settings that we did in the previous case of small output noise.

Fig 1. shows the comparison among four methods in term prediction errors (RMSE) while Fig 2. shows the input recovery error inccured by our method and MCMC. 

\begin{figure}[!htbp] 
	\begin{subfigure}[b]{0.5\linewidth}
		\centering
		\includegraphics[width=0.95\linewidth]{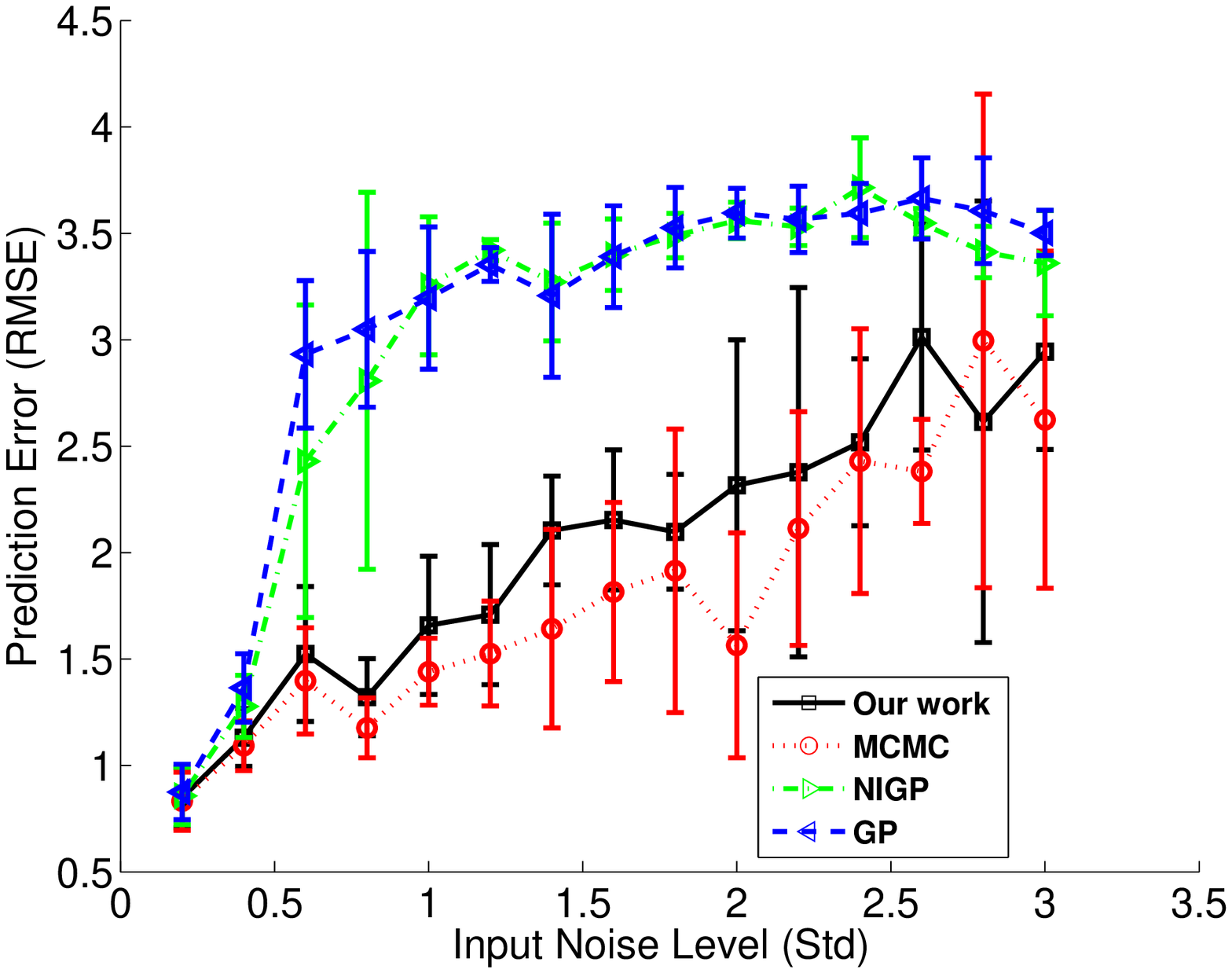} 
		\caption{ $ f_1(\tau)=5sin(\tau) $} 
		\label{fig7:a} 
		\vspace{2ex}
	\end{subfigure}
	\begin{subfigure}[b]{0.5\linewidth}
		\centering
		\includegraphics[width=0.95\linewidth]{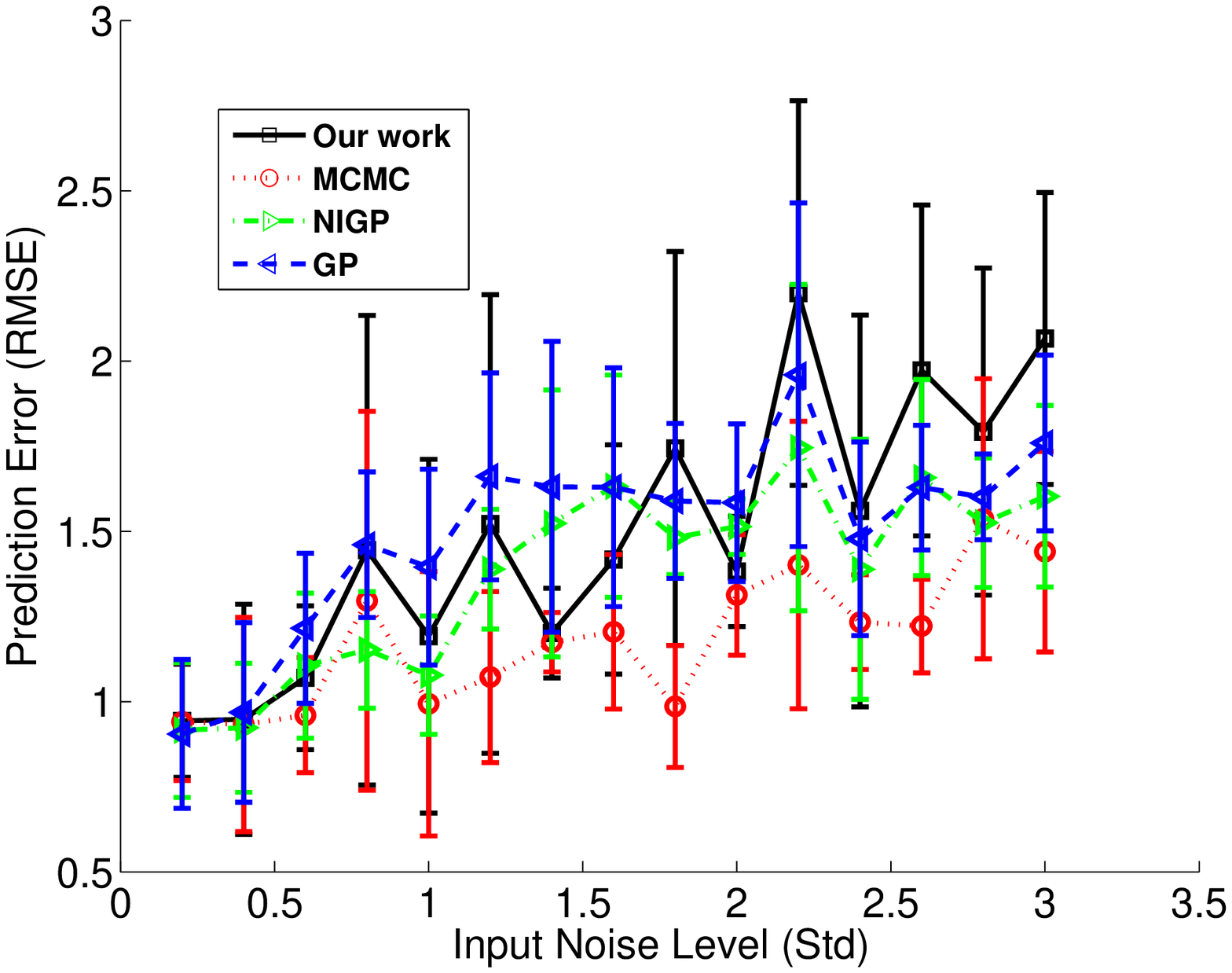} 
		\caption{$ f_2(\tau)=1.147 e^{-0.2 \tau}sin(\tau)$} 
		\label{fig7:b} 
		\vspace{2ex}
		
	\end{subfigure} 
	\begin{subfigure}[b]{0.5\linewidth}
		\centering
		\includegraphics[width=0.95\linewidth]{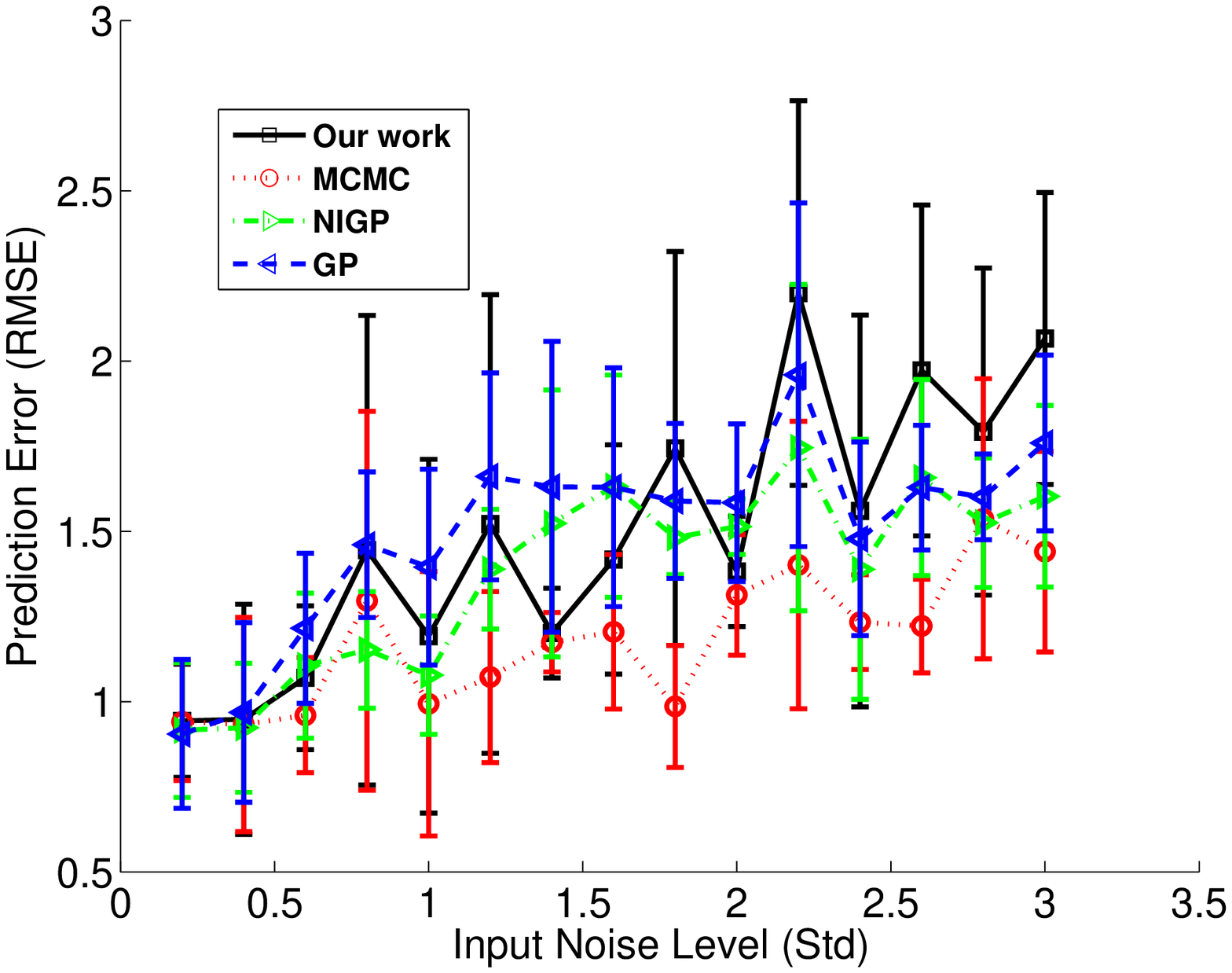} 
		\caption{$ f_3(\tau)=0.97 tan(0.15 \tau) sin(\tau)$} 
		\label{fig7:c} 
		\vspace{2ex}
	\end{subfigure}
	\begin{subfigure}[b]{0.5\linewidth}
		\centering
		\includegraphics[width=0.95\linewidth]{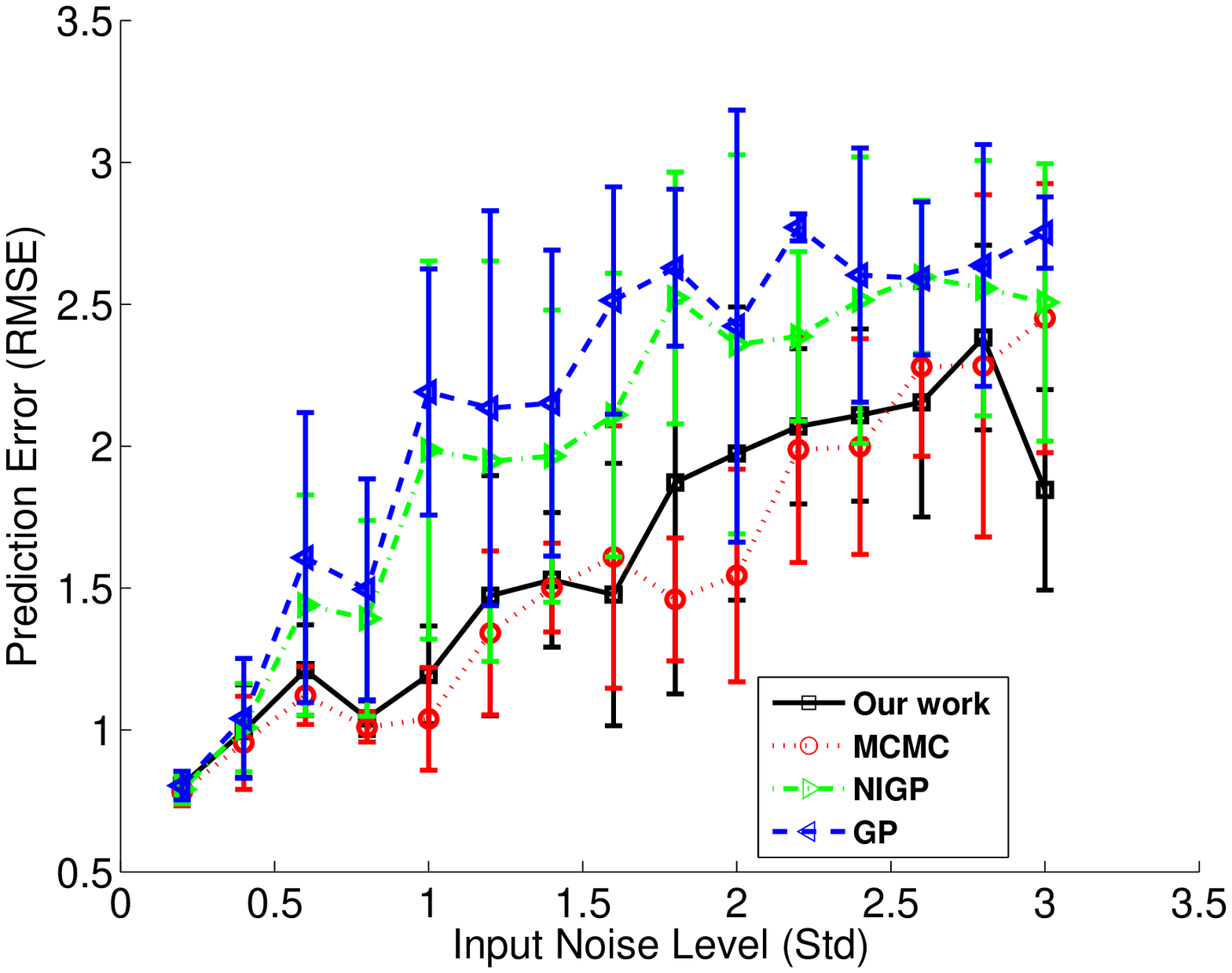} 
		\caption{$f_4(\tau)=0.055 \tau^2 tanh (cos(\tau))$ } 
		\label{fig7:c} 
		\vspace{2ex}
	\end{subfigure}
	
	\begin{subfigure}[b]{0.5\linewidth}
		\centering
		\includegraphics[width=0.95\linewidth]{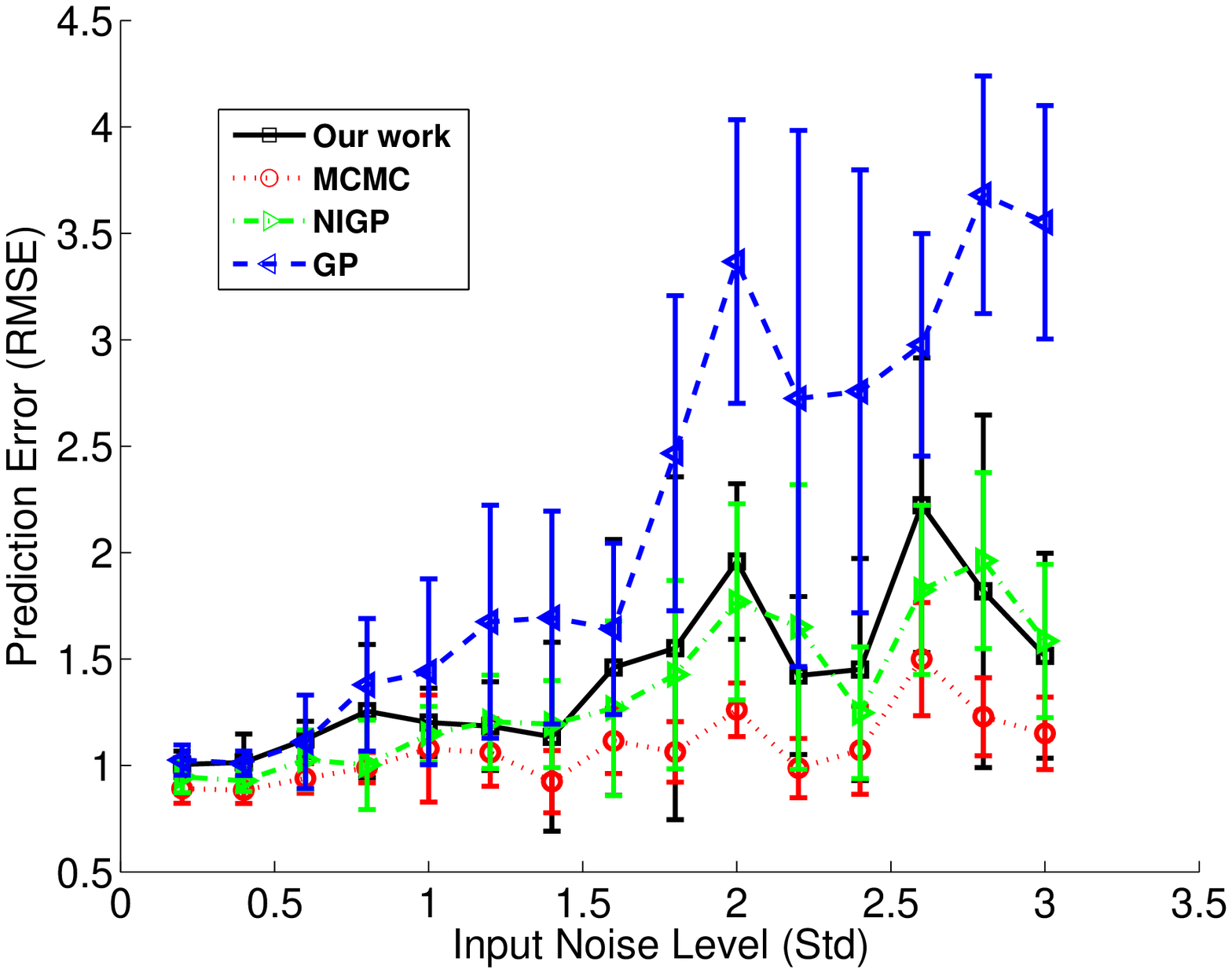} 
		\caption{$f_5(\tau)=1.76 log( \tau^2 (sin (2\tau)+1)+1)$} 
		\label{fig7:a} 
		\vspace{2ex}
	\end{subfigure}
	\caption{ Comparison among four methods: our work, MCMC, NIGP and GP based on prediction errors when the ouput noise is large. In this case, our method is still better than usual GP. However, it can not outperform NIGP as in the case of small output noise. }
	\label{fig7} 
\end{figure}

\begin{figure}[!htbp] 
	\begin{subfigure}[b]{0.5\linewidth}
		\centering
		\includegraphics[width=0.95\linewidth]{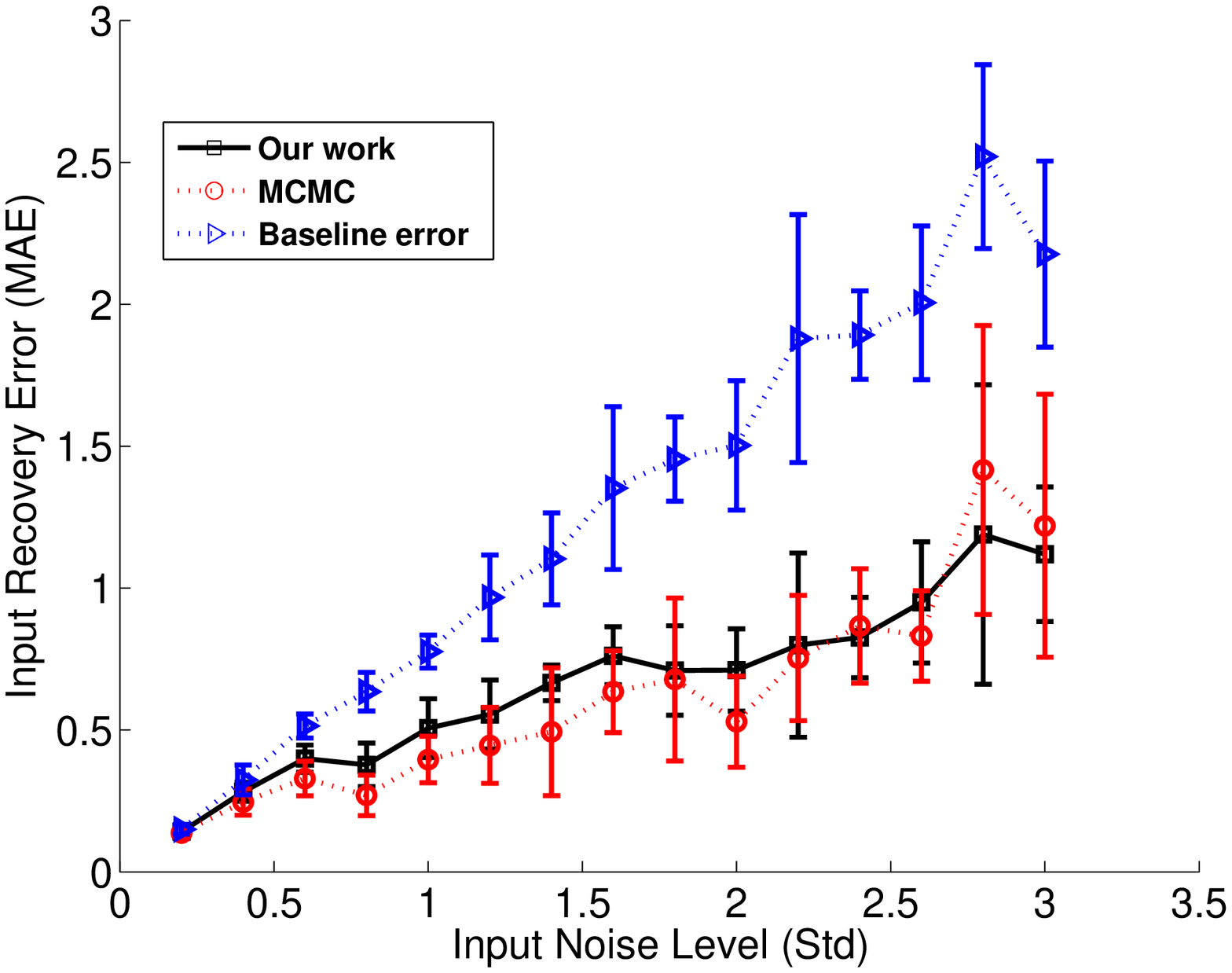} 
		\caption{ $ f_1(\tau)=5sin(\tau) $} 
		\label{fig7:a} 
		\vspace{2ex}
	\end{subfigure}
	\begin{subfigure}[b]{0.5\linewidth}
		\centering
		\includegraphics[width=0.95\linewidth]{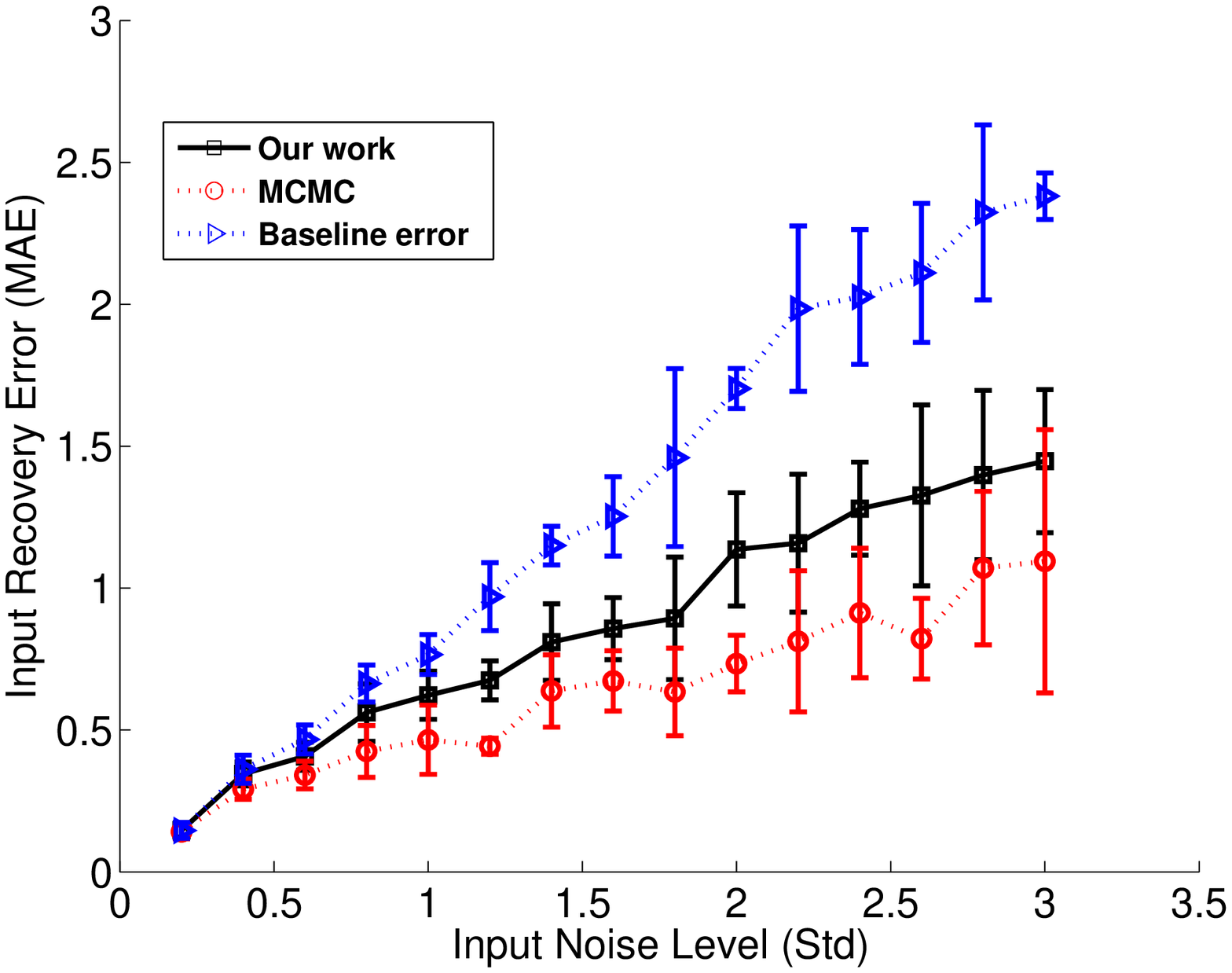} 
		\caption{$ f_2(\tau)=1.147 e^{-0.2 \tau}sin(\tau)$} 
		\label{fig7:b} 
		\vspace{2ex}
		
	\end{subfigure} 
	\begin{subfigure}[b]{0.5\linewidth}
		\centering
		\includegraphics[width=0.95\linewidth]{Lf3.eps} 
		\caption{$ f_3(\tau)=0.97 tan(0.15 \tau) sin(\tau)$} 
		\label{fig7:c} 
		\vspace{2ex}
	\end{subfigure}
	\begin{subfigure}[b]{0.5\linewidth}
		\centering
		\includegraphics[width=0.95\linewidth]{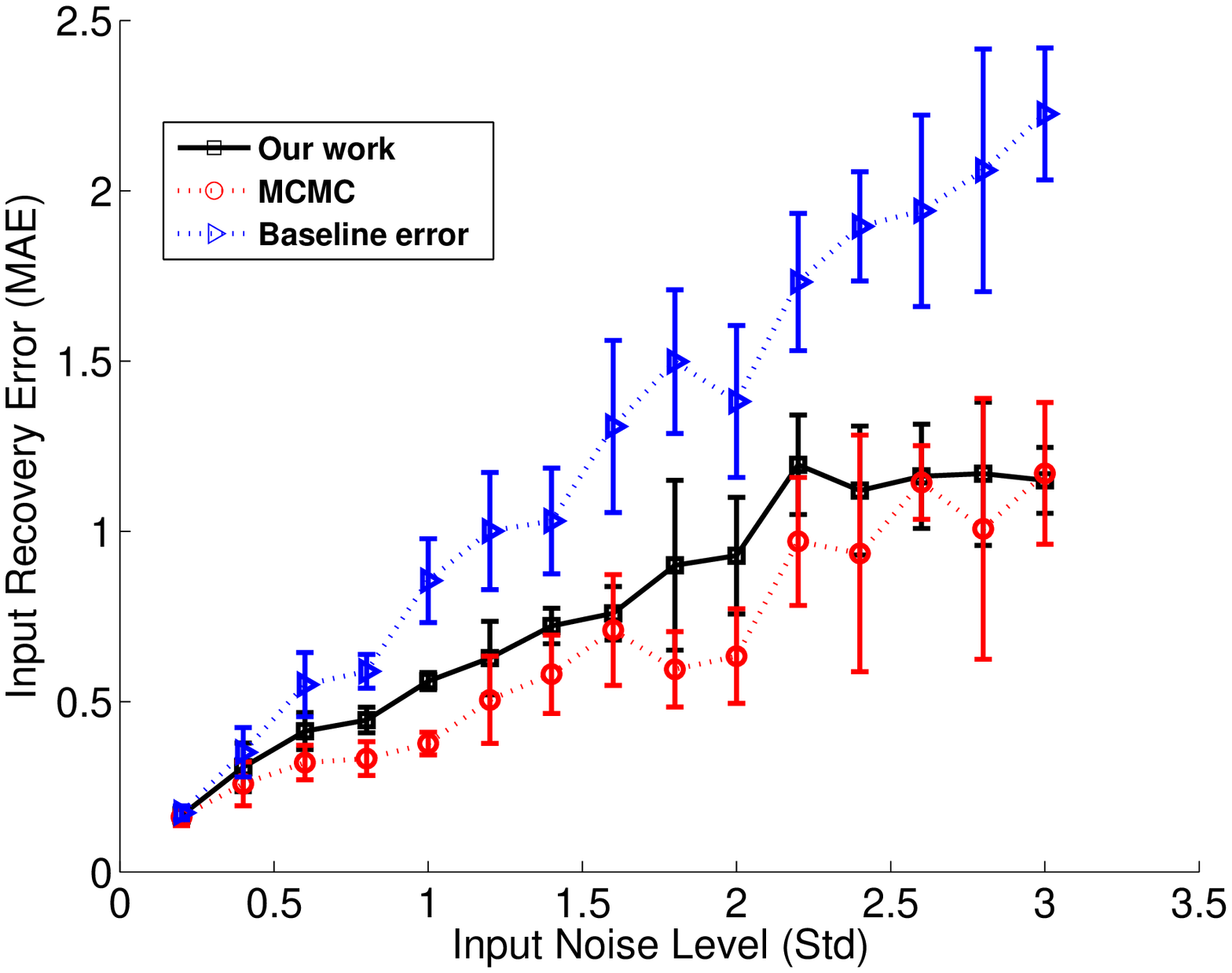} 
		\caption{$f_4(\tau)=0.055 \tau^2 tanh (cos(\tau))$ } 
		\label{fig7:c} 
		\vspace{2ex}
	\end{subfigure}
	
	\begin{subfigure}[b]{0.5\linewidth}
		\centering
		\includegraphics[width=0.95\linewidth]{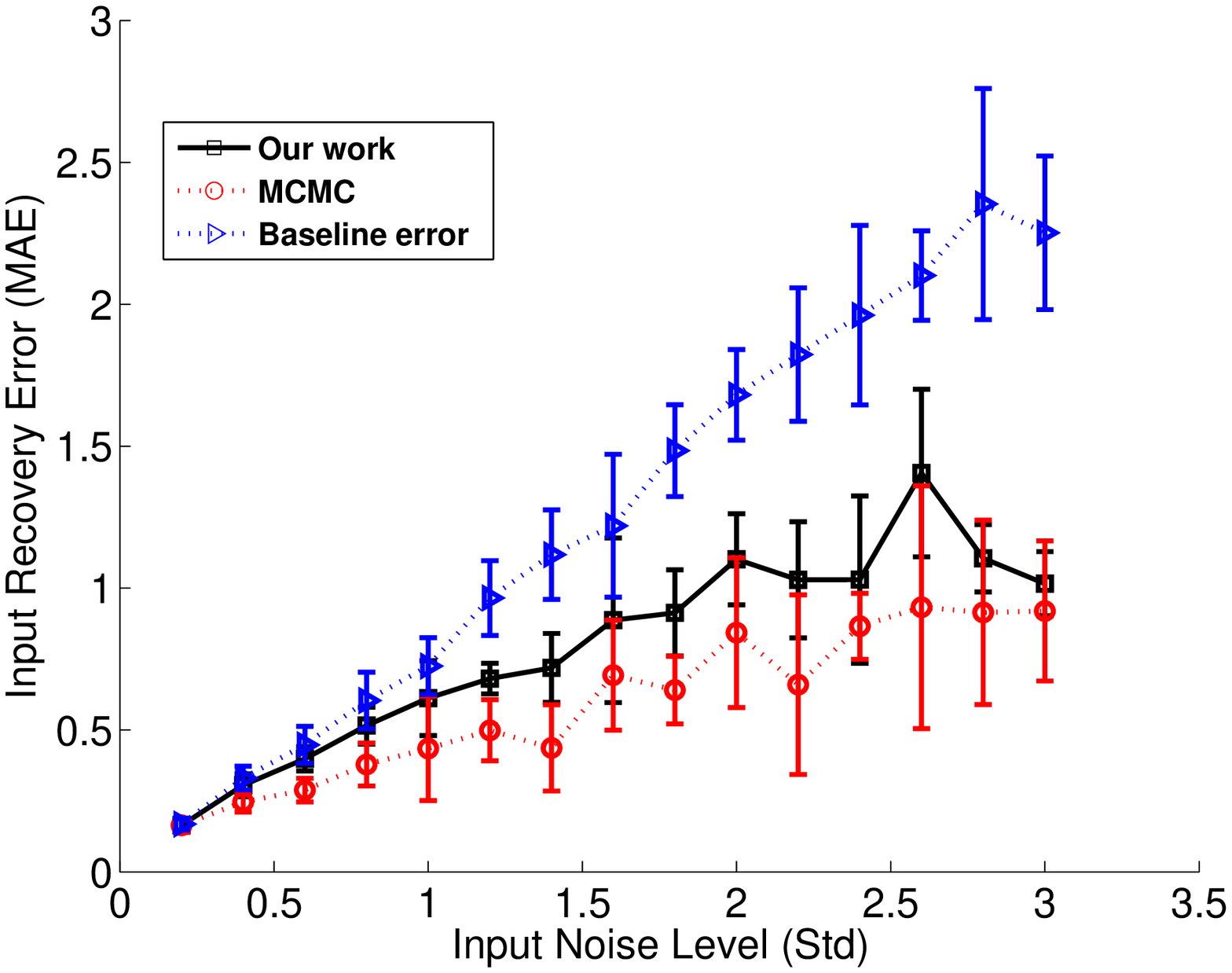} 
		\caption{$f_5(\tau)=1.76 log( \tau^2 (sin (2\tau)+1)+1)$} 
		\label{fig7:a} 
		\vspace{2ex}
	\end{subfigure}
	\caption{Comparison between our work with MCMC in term the ability to recover the true inputs when output noise is large. The baseline error measures the distance from noisy inputs $t$ to the true inputs $\tau$ }
	\label{fig7} 
\end{figure}